\documentclass[letterpaper, 10 pt, conference]{ieeeconf}  

\IEEEoverridecommandlockouts                              
                                                          
\IEEEoverridecommandlockouts                            
\usepackage{breqn}

\usepackage{color}
\usepackage{colortbl}
\usepackage{graphicx} 
\usepackage[american]{babel}
\usepackage{subfigure}		
\usepackage{balance}
\usepackage{cite}
\usepackage{xcolor}
\usepackage{hyperref}
\usepackage{amsmath}
\usepackage{bbding}
\usepackage{bm}
\usepackage{fontawesome}
\usepackage{nicematrix}
\usepackage{array}
\usepackage{dblfloatfix}    
\usepackage{stackengine}

\usepackage{amssymb}

\usepackage{algorithm}
\usepackage{algpseudocode}
\usepackage{multirow} 

\title{\LARGE \bf
CineTransfer: 
Controlling a Robot to Imitate 
Cinematographic Style from a Single Example
}

\author{\centering Pablo Pueyo, Eduardo Montijano, Ana C. Murillo and Mac Schwager
\thanks{This work was supported by a DGA scholarship and  project  T45\_23R; Spanish project PID2021-125514NB-I00, funded by MCIN/AEI/10.13039/501100011033, by ERDF A way of making Europe and by the European Union NextGenerationEU/PRTR; NSF grants CNS-1330008 and IIS-1646921; ONR grant N00014-18-1-2830.}
\thanks{P. Pueyo, E. Montijano and A. C. Murillo are with DIIS and I3A, Universidad de Zaragoza, Spain 
\texttt{\small \{ppueyor, emonti, acm\}@unizar.es}}
\thanks{M. Schwager is with Dept. of Aeronautics and Astronautics, Stanford University, USA
\texttt{\small \{schwager\}@stanford.edu}}
}

\begin{document}

\maketitle
\thispagestyle{empty}
\pagestyle{empty}

\begin{abstract}
This work presents CineTransfer, an algorithmic framework that drives a robot to record a video sequence that mimics the cinematographic style of an input video. We propose features that abstract the \emph{aesthetic style} of the input video, so the robot can transfer this style to a scene with visual details that are significantly different from the input video.  The framework builds upon CineMPC, a tool that allows users to control cinematographic features, like subjects' position on the image and the depth of field, by manipulating the intrinsics and extrinsics of a cinematographic camera.
However, CineMPC requires a human expert to specify the desired style of the shot (composition, camera motion, zoom, focus, etc). CineTransfer bridges this gap, aiming a fully autonomous cinematographic platform. The user chooses a single input video as a style guide. CineTransfer extracts and optimizes two important style features, the composition of the subject in the image and the scene depth of field, and provides instructions for CineMPC to control the robot to record an output sequence that matches these features as closely as possible.
In contrast with other style transfer methods, our approach is a lightweight and portable framework which does not require deep network training or extensive datasets. 
Experiments with real and simulated videos demonstrate the system’s ability to analyze and transfer style between recordings, and are available in the supplementary video\footnote{\label{footnote_1}\url{https://youtu.be/_QzNz5WUtpk}}.

\end{abstract}


\section{Introduction}
\label{sec_intro}



 The use of robots equipped with cameras for cinematographic and recreational purposes has grown significantly in recent years. This has resulted in a surge of movies featuring scenes captured by camera-equipped drones or automatic cranes, as well as an increase in amateur users utilizing mobile cameras to take photos and videos. However, operating a robot can be challenging for individuals who are not familiar with the cinematographic or robotic world. 

To overcome this, current solutions propose control techniques that enable mobile robots to autonomously record scenes~\cite{bonatti2020autonomous, alcantara2020autonomous, pueyo2022cinempc}. These solutions require users to provide artistic or technical guidelines for the desired footage, after which the robot follows trajectories to capture the scene as instructed. 
In particular, our previous method, CineMPC~\cite{pueyo2022cinempc}, uses a Model Predictive Control (MPC) framework to control both the extrinsics (camera pose) and intrinsics (focal length, aperture, focus distance) of the camera. CineMPC requires human user input to specify the desired scene composition and focus characteristics of the shot, which are encoded as instructions of an MPC optimization problem that outputs a trajectory for the drone and camera parameters. 
While these solutions, in particular CineMPC, have made significant progress in the field, they 
fail to address a crucial aspect: intuitive human-robot interaction for easy specification of recording instructions.
\begin{figure}[!t]
\centering
\begin{tabular}{c}
    \includegraphics[width=0.99\columnwidth]{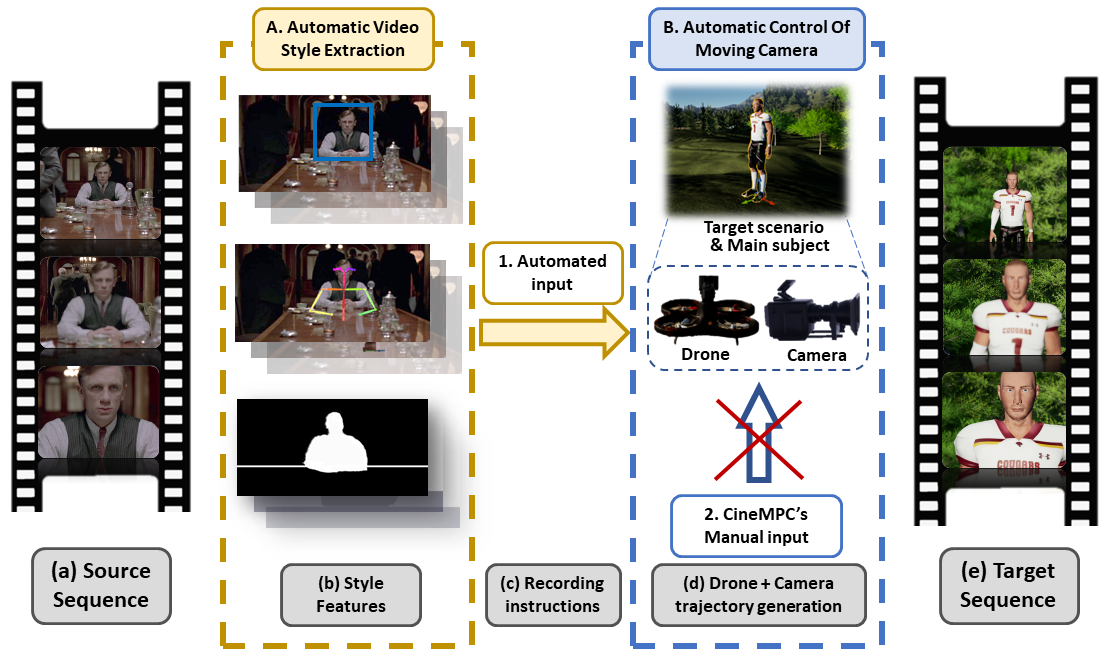}
\end{tabular}
\caption{\footnotesize{\textbf{CineTransfer} automatically extracts the style of a source input video \textbf{(a)} and moves a robot with a camera to record a video that mimics the source style in a different scenario \textbf{(e)}. We first \textbf{(A)} extract style features from the source video 
and solve an optimization problem to ensure the continuity and quality of the features. The style is defined by the position of the main subject, his/her body joints, and the focus of the scene \textbf{(b)}. From these features, we automatically create instructions for an MPC controller to transfer the style \textbf{(c)}. The second part \textbf{(B)}, built upon CineMPC, solves the MPC problem to define the trajectory \textbf{(d)} of a mobile robot holding a cinematographic camera to record a new video according to the instructions. CineTransfer replaces manual recording instructions \textbf{(2)} required by CineMPC with automated recording instructions that represent the source style of the input video \textbf{(1)}.
}
}
\label{fig:main}
\end{figure}



To address this issue, we propose CineTransfer, an approach that significantly improves the level of autonomy for robotic cinematography. With CineTransfer, users only need to choose a video clip with the desired cinematographic composition and style,  and  the platform recreates the style of this clip in the user's target scenario. The visual details of the two scenarios can be quite different, while the aesthetic style of the shots will still be similar.  For example, users can select as a source a sequence from a movie recorded by expert cinematographers. CineTransfer automatically extracts and optimizes the features that define the style from the sequence and converts them to recording instructions for CineMPC, irrespective of the details of where and how the sequence was recorded.


Our approach, summarized in Fig.~\ref{fig:main}, analyzes two of the most important cinematographic features for each source sequence frame: the main subject localization and his/her body joints in the image, and the parts of the scene that appear in focus or out of focus. Our proposed optimization step helps to reduce the noise from these measures, ensuring the continuity between the features on subsequent frames. The results are collected and presented to CineMPC as recording instructions to mimic the source sequence style in a target scenario where the subject and background might be different.  
Our results demonstrate the approach ability to extract and reproduce the style features from source videos in different scenarios.

\section{Related work}
\label{sec_related}
%

Controlling a robot using images has attracted the interest of the research community for years.
For instance, solutions on visual servoing~\cite{janabi2010comparison}, move a camera to align the captured image with a desired image. 
There are also different extensions to sequences of images for robot navigation~\cite{chen2009qualitative, matsumoto1996visual}. 
More recent works incorporate machine-learning to solve the task \cite{paradis2021intermittent}. 
In contrast to our work, these works 
must be run on the same scene, i.e., there is no cross-scene generalization.



In the field of cinematography, there exist solutions for ``imitation filming'', whose aim is to control a robotic camera to reproduce a video sequence. The pipeline described in  \cite{dang2020imitation} trains a network with video demonstrations to generate a trajectory that optimizes the viewpoints with respect to the subject. 
The same authors extend the work in \cite{dang2022path} with an algorithm that classifies and recreates the style that defines the relative pose between camera and subject from an input sequence, using multiple layers of imitation learning. 
 The work in \cite{jiang2021camera} presents a framework that uses machine learning to move a drone to reproduce the style of a reference movie clip, passing through a set of given camera viewpoints. 
 The paper \cite{huang2019learningtwo} considers an imitation learning framework where a network predicts camera motions using a dataset of human-based videos recorded by professional cinematographers.
Similarly, imitation learning is used in \cite{huang2019learning} to extract the features of a shot that a control algorithm uses as a reference to move a drone to record a similar scene. 
More recently~\cite{huang2021one}, 
two networks, a camera motion predictor and a style feature detector, classify the input video in one out of five possible human-centered styles and reproduce it. 
These methods require large datasets and training data to achieve convincing results.
Differently, CineTransfer is a lightweight and portable approach which does not require any training or extensive datasets to obtain similar results. 

\section{System Overview}
\label{sec_problem_formulation}
The overall objective of CineTransfer is to record a cinematographic sequence that reproduces the style of a different input video sequence. Figure  \ref{fig:main} summarizes the whole process, which is divided into two separate modules.
In the first module, called \textbf{Automatic Video Style Extraction} (Fig.\ref{fig:main}-A), the framework extracts and optimizes the style features concerning the depth of field of the scene and the framing of the subject from the source video, and composes the instructions for the second module of the framework. 
In the second module, called \textbf{Automatic Control of Moving Camera} (Fig.\ref{fig:main}-B), the recording instructions that define the style of the source video sequence are transformed into commands that are sent to a cinematographic camera, satisfying the instructions.
The following sections describe the modules in detail. 
We describe first the second part to introduce the features that need to be extracted from the source sequence to perform the control.

 \section{Automatic Control of Moving Camera}
\label{sec_cinempc}
This module of the framework is a variation of our previous solution~\cite{pueyo2022cinempc}. CineMPC is an algorithm that solves a non-linear optimization problem inside a Model Predictive Control (MPC) framework to autonomously control a camera together with a mobile robot for cinematographic purposes. 

The algorithm calculates the trajectory of the extrinsic and intrinsic parameters that the robot and camera need to execute in order to satisfy the recording instructions. Among all the cinematographic features controllable by CineMPC, we focus this work on the desired position of the elements of the scene in the image and the desired Depth of Field (DoF), or focus of the image.
CineMPC solves the following problem,
\begin{equation}
\label{Eq:Cinempc}
\begin{aligned}
\min_{\substack{\mathbf{u}_{d,k_0}..\mathbf{u}_{d,k_0+N}\\ \mathbf{u}_{c,k_0}..\mathbf{u}_{c,k_0+N}}} \quad & \sum_{k=k_0}^{k_{0}+N}
    J_{im,k}+{J_{DoF,k}},\\
\textrm{s.t.} \quad & \mathcal{D}_k\ \hbox{and}\ \mathcal{C}_k.  \\
\end{aligned}
\end{equation}
This equation is used to calculate the next $N$ drone and camera actions, $\mathbf{u}_{d,k}$ and $\mathbf{u}_{c,k}$, respectively, at a giving time $k_0$. The problem is determined by the dynamics and constraints of the drone and camera, that are respectively stored in the sets $\mathcal{D}_k$ and $\mathcal{C}_k$. 
For a more detailed description of the complete optimization, we refer the reader to~\cite{pueyo2022cinempc}.

The relevant information to understand CineTransfer is encoded in the two cost terms that appear in~\eqref{Eq:Cinempc}.
The first term, $J_{im}$, controls the \textbf{position of the elements on the image}. The cost is composed of the actual and desired image position of element $e$ of the scene ($\textbf{im}_{e,k}$ and $\textbf{im}_{e,k}^*$) and the corresponding weight, assuming that there are $E$ elements of interest,
\begin{equation}
J_{im,k} = \sum_{e=1}^{E}{\|\textbf{im}_{e,k} - \textbf{im}_{e,k}^{*}\|^2}.
    \label{eq:im-cost}
\end{equation}

The second term of the cost function, $J_{DoF}$, controls the \textbf{depth of field}, which is the area of the scene is in focus in the image. 
The cost is composed by the current and desired near ($D_n$ and $D_n^*$) and far distances ($D_f$ and $D_f^*$), which are the distances from the camera that delimit the depth of field, 
\begin{equation}
J_{DoF,k} = \left(D_{n,k} - D_{n,k}^{*}\right)^2 + \left(D_{f,k} - D_{f,k}^{*}\right)^2.
    \label{eq:dof-cost}
\end{equation}
 
In order to work properly, CineMPC requires an expert that provides the values of all the desired quantities, $E$, $\textbf{im}_{e,k}^{*}$, $D_{n,k}^{*}$, and $D_{f,k}^{*}$. 
The objective of the Automatic Video Style Extraction module is precisely to extract these parameters without any human intervention from the source sequence. 
The next section describes this process in detail.
 


 \section{Automatic Video Style Extraction}
\label{sec_agents}
\begin{figure}[!t]
\centering
    \includegraphics[width=0.9\columnwidth,height = 6cm]{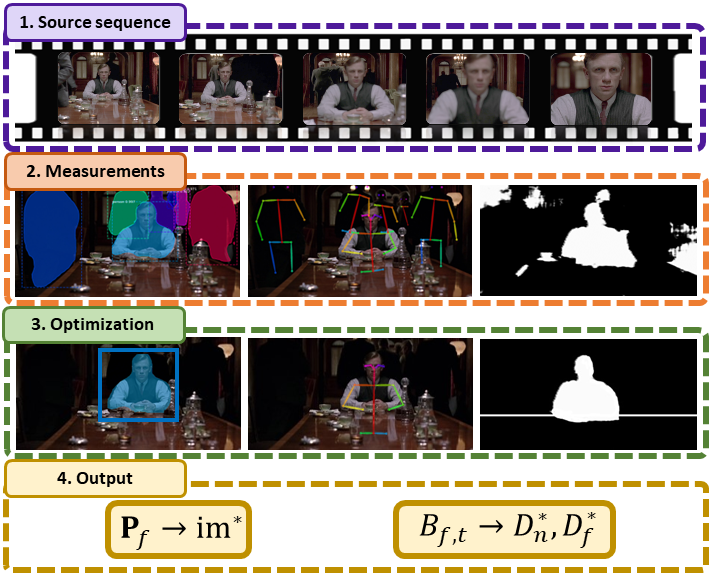} 
 \caption{\textbf {Automatic Video Style Extraction Pipeline.} Steps of the pipeline executed sequentially and ordered in time from top to bottom.\textit{ (1) Input source sequence.} \textit{(2) Feature measurements obtained on a single frame}. (left)  segmentation masks and bounding boxes of the subjects; (center) body joints and skeleton of the subjects; (right) estimation of focus at different areas of the image. White regions represent the focused areas of the scene, and black regions represent the blurry areas. \textit{(3) Optimization step}. Optimization libraries ensure the quality and continuity of the measurement.
\textit{ (4) Output.} 
Recording instructions of the second module (see Fig.~\ref{fig:main}). }
\label{fig:pipeline}
\end{figure}

To estimate the required reference features from the source video,
the framework uses existing models to extract a variety of measurements of interest. Then, it uses several optimization procedures over all of them to filter out noise and ensure continuity during the whole sequence. Finally, the method converts the output of the optimization to the desired recording instructions that are fed into CineMPC. The next subsections describe all the steps in detail. The diagram of Fig.~\ref{fig:pipeline} represents the flow of this pipeline.

\subsection{Input (Fig. \ref{fig:pipeline}-1)}
The input is the source video sequence. We extract the frames of the sequence and analyze them one by one. Each frame of the sequence is denoted as $f$. 

\subsection{Measurements. (Fig. \ref{fig:pipeline}-2)}
To detect the position of the main subject and his/her body joints, we run two preexisting deep-learning libraries for each frame; 
Mask-RCNN \cite{matterport_maskrcnn_2017} and OpenPose \cite{openpose2019} due to their demonstrated efficiency~\cite{li2019crowdpose,wang2021urbanpose}.

 \textbf{Mask-RCNN} detects multiple objects on single images and provides their class (e.g., person, car, etc.).
For each person detected in each frame, denoted by $m \in \{1,...,M_f\}$, Mask-RCNN also returns a confidence, $c_{f,m}\in \mathbb{R}$, the pixels of the image where is placed (the mask), $\mathbf{m}_{f,m}$, and a bounding box, 
$\mathbf{r}_{f,m} \in \mathbb{R}^{4}$ (left image in Fig.~\ref{fig:pipeline}-2).
\textbf{OpenPose} detects people in the frame and extracts their relevant body joints.
The body joints of person $p$ are grouped in the vector $\mathbf{p}_{f,p}$. For each body joint, OpenPose extracts its position on the frame 
$\in \mathbb{R}^{2}$ and the confidence on the measure $q_{f,p} \in \mathbb{R}$ (middle image in Fig.~\ref{fig:pipeline}-2).
To measure the focus we use \textbf{EFENet}~\cite{zhao2021defocus}, an off-the-shelf library that builds a map of Boolean values representing if each pixel of the frame is in focus or not (right image in Fig.~\ref{fig:pipeline}-2).

\subsection{Optimization. (Fig. \ref{fig:pipeline}-3)}
In this step, we solve  different optimization problems to filter the noise and ensure smoothness of the features over the sequence. First, we identify the main subject, then, we optimize the location of his/her body joints in the image and, finally, we compute the focus style of the scene.

\subsubsection{Detection of main subject}
The objective is to localize the main subject in all the frames by analyzing the quality and continuity of the measurements. The main subject is the subject that appears in the majority of the frames. 
Given the bounding boxes of the visible subjects in each frame,
$\mathbf{r}_{f,m}$, and their an associate confidence, $c_{f,m}$, 
the optimization problem 
\begin{equation*}
\begin{aligned}
\min_{\substack{\mathbf{R}_{f}\\{\alpha_{f,m}}}} \; &\sum_{f=1}^{F}\Bigl( \|\mathbf{R}_{f} - \mathbf{R}_{f-1}\|^2 + \sum_{m=1}^{M_f} c_{f,m}\alpha_{f,m}\|\mathbf{R}_{f} - \mathbf{r}_{f,m}\|^2\Bigr),\\
&\textrm{s.t.}
 \quad \mathbf{R}_{f} \in \mathcal{I},\ \sum_{m=1}^{M_f}{\alpha_{f,m}} = 1,  \ 0 < \alpha_{f,m}< 1, 
\label{eq:cost-pixels}
\end{aligned}
\end{equation*}
returns a bounding box in each frame, $\mathbf{R}_{f}\in \mathbb{R}^{4},$ where the main subject should appear.
The parameters $F$ and $M_f$, represent the total number of frames and the number of subjects found in the frame, respectively. 
The decision variable $\alpha_{f,m}$ is an auxiliary variable to make sure that only one bounding box out of the $M_f$ measured is considered in each frame.
The constraint $\mathbf{R}_{f} \in \mathcal{I}$ is used to force all the bounding boxes to be visible inside the image limits.

\subsubsection{Detection of body joints of the main subject}
To determine the list of points of interest of the main subject of a frame, $\mathbf{P}_{f}$,
we first find his/her mask, $\mathbf{M}_{f}$. To do so, we select the mask that contains the most pixels inside of the bounding box of the main subject, 
\begin{equation}
\label{Eq:optimalMask}
    \mathbf{M}_{f} =  \arg\max_{\mathbf{m}_{f,m}}\Bigl( g(\mathbf{m}_{f,m}, \mathbf{R}_{f}) \Bigr), 
 \end{equation}
where $g(\mathbf{m}_{f,m},\mathbf{R}_{f})$ is the number of pixels of the mask $\mathbf{m}_{f,m}$ that are inside the bounding box $\mathbf{R}_{f}$.

The list of body joints is obtained in a similar fashion, searching for the person with more joints inside $\mathbf{M}_{f}$,
\begin{equation}
    \mathbf{p}_{f} =  \arg \max_{\mathbf{p}_{f,p}}\Bigl( h(\mathbf{p}_{f,p}, \mathbf{M}_{f}) \Bigr), 
 \end{equation}
where $h(\mathbf{p}_{f,p},\mathbf{M}_f)$ is the number of body joints of the list $\mathbf{p}_{f,p}$ that are inside the mask $\mathbf{M}_f$.

Then, another optimization problem favors the continuity and quality of each component of the list of body joints, grouped in $\mathbf{P}_{f}$,
\begin{equation}
\begin{aligned}
\min_{\substack{\mathbf{P}_{f}}} & \; \sum_{f=1}^{F}\Bigl( \|\mathbf{p}_{f} - \mathbf{p}_{f-1}\|^2 + \sum_{p=1}^{P_f} q_{f,p}\|\mathbf{P}_{f} - \mathbf{p}_{f,p}\|^2\Bigr)\\
&\textrm{s.t.}
 \quad \mathbf{P}_{f} \in \mathcal{I} 
\label{eq:cost-pixels}
\end{aligned}, 
\end{equation}
where $P_f$ is the number of people found in frame $f$.

\subsubsection{Focus of scene}
The focus style is determined by whether the foreground, 
the main subject, and the background 
appear in focus in the image or not.
We use the optimized mask from Eq.~\eqref{Eq:optimalMask}, $\mathbf{M}_{f}$, to know the regions of the blurry map that belong to the background, main subject, and foreground.
For each $t \in $ \textit{\{foreground, main subject, background\}}, we calculate the percentage of pixels that are in focus,
\begin{equation}
   b_{f,t}=\frac{\beta(\mathbf{M}_{f,t})}{ |\mathbf{M}_{f,t}|},
\end{equation}
where the function $\beta(\cdot)$ returns the total number of pixels of the mask that are in focus.

The last optimization problem is used to smooth out the preliminary output, avoiding misleading measurements. This problem favors continuity of the percentage of focused points of the subject in the sequence (first term) but trying to make it close to zero or one (second term),
\begin{dmath}
\min_{\substack{B_{f,t}}} \quad  \sum_{f=1}^{F}  \biggl( \Bigl(B_{f,t} - B_{f-1,t}\Bigr)^2 + \Bigl( B_{f,t}  (1-2b_{f,t})\Bigr)^2  \biggr), \\
\textrm{s.t.} \quad  {0<B_{f,t}<1}.
\end{dmath}
The output $B_{f,t}$ represents the percentage of focused pixels for the region $t$ within a determinate frame $f$.
If it is above a threshold, $\theta$, we set
the region to appear in focus.

\subsection{Output. (Fig. \ref{fig:pipeline}-4)}
\label{sec:output}
The last step transforms the results into recording instructions for the Automatic Control of Moving Camera module. 
 
 The \textbf{framing of the main subject} is determined by the position of the relevant body joints $p$ of the main subject, e.g., control the shoulders of the main subject horizontally, which are extracted and reproduced from the source sequence, and stored in the vector $\mathbf{P}_f$ for each frame. As there exist multiple screen sensor sizes, e.g. 1980x920 px, 960x540 px, the body joints of $\mathbf{P}_{f}$ are normalized from 0 to 1. Then, each camera can place the points in a position of the frame proportional to its sensor size. 
The positions of the elements of the image that are controlled in Eq.~\eqref{eq:im-cost} are set according to the relevant body joints of the main subject,
  \begin{equation}
  \mathbf{im^*}_{e,k}=\mathbf{P}_{p,f},\ \ E = |\mathbf{P}_{f}|.
\end{equation} 
Each time step $k$ is set to the duration of the frame $f$.

To represent the \textbf{focus of the scene} the optimization step calculates three boolean values for each frame. 
 The variable $B_{f,0}$ is boolean and true if the foreground is in focus. Analogously, $B_{f,1}$ and $B_{f,2}$ represent the focus of the main subject and background respectively. 
 Let $d_k$ be the distance from the camera to the main subject at a time step $k$. The desired near and far distances in Eq. \eqref{eq:dof-cost} are set as
\begin{equation*}
  D_{n,k}^*=\left\{
  \begin{array}{@{}ll@{}}
    0, & \text{if}\ B_{f,0} \\
    d_k-\mu & \text{elsif}\ B_{f,1} \\
    d_k+\mu & \text{elsif}\ B_{f,2} \\
    \infty & \text{else}
  \end{array}\right.,
  D_{f,k}^*=\left\{
  \begin{array}{@{}ll@{}}
    \infty, & \text{if}\ B_{f,2} \\
    d_k+\mu & \text{elsif} \ B_{f,1} \\
    d_k-\mu & \text{elsif} \ B_{f,0} \\
    -\infty & \text{else}  \\
  \end{array}\right.,
\end{equation*} 
where $\mu$ determines the size of the region of the depth of field ($2 \mu$ if $D_n^*$ and $D_f^*$ are controlled) and how sharp the focus of the elements is, and $k$ is adapted to the duration of the frame.

\section{Experimental Validation}
\label{sec_experiments}
This section evaluates CineTransfer with different source sequences and target environments.
Recordings take place in a virtual environment scenario using the photo-realistic simulator CinemAirSim \cite{pueyo2020cinemairsim}. 
For more detailed visual results, we refer the readers to the supplementary video.

 \subsection{Data used}
We selected two \textit{source sequences} (included in the supplementary video) that include the style aspects to be transferred.
\begin{itemize}
    \item Source sequence 1 (S1): Sequence taken from a real movie (Road To Perdition-2002 by Sam Mendes). The main subject remains static while sitting at a table, but the framing and the depth of field change.
    \item Source sequence 2 (S2): Simulated sequence with ground truth information for quantitative evaluation. The main subject 
    is walking, and the focus and framing vary significantly through the scene.
\end{itemize}
For the \textit{target scenarios}, i.e., the 3D scene where CineMPC will record the new sequence, we selected three simulation environments.
\begin{itemize}
    \item Target scenario 1 (T1): Reproduction of the scenario of S1, with a very similar main subject and setup. 
    \item Target scenario 2 (T2): Same scenario as  S2, for qualitative comparisons.
    \item Target scenario 3 (T3): Scenario very different from source sequences. A football player standing in an open mountain landscape, surrounded by trees.
\end{itemize}

 \subsection{Implementation details}
 All the experiments are run in Ubuntu 20 on an AMD Ryzen 9 5900X 12-Core CPU equipped with 32 Gb of RAM and an NVidia GeForce GTX 3090 Ti. CineTransfer is implemented in Python. It makes calls to the three aforementioned libraries 
 (Mask-RCNN~\cite{matterport_maskrcnn_2017}, OpenPose ~\cite{openpose2019} and EFENet~\cite{zhao2021defocus}), and collects the results. The Python optimization library Gekko \cite{gekko} optimizes these results, which are stored in files with JSON format. Then, CineMPC, implemented in ROS and C++, has been adapted to read the recording instructions from the JSON files and use them as control instructions.

  \subsection{Quality of style features extracted}

  This subsection analyzes the quality of the features extracted by CineTransfer to describe the style from the source videos. 
  We compare the optimized and raw feature values obtained by our framework with the ground truth.

\subsubsection{\textbf{Detection of main subject position and mask}}

\begin{figure}[!tbh]
\centering
\begin{tabular}{cc}   
    \includegraphics[width=0.38\columnwidth]{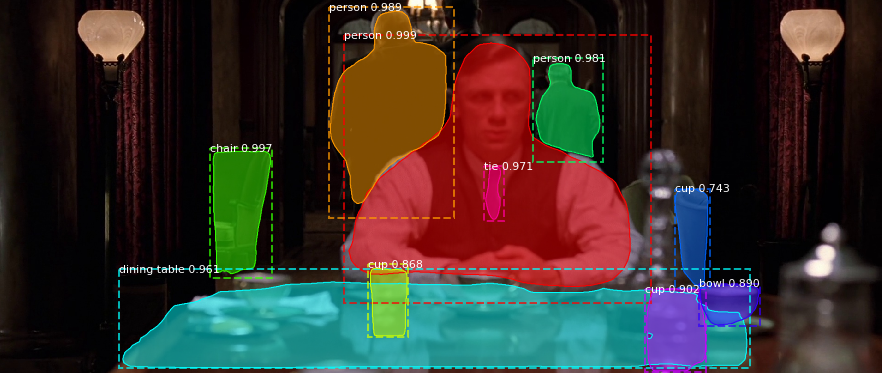}
    &    \includegraphics[width=0.38\columnwidth]{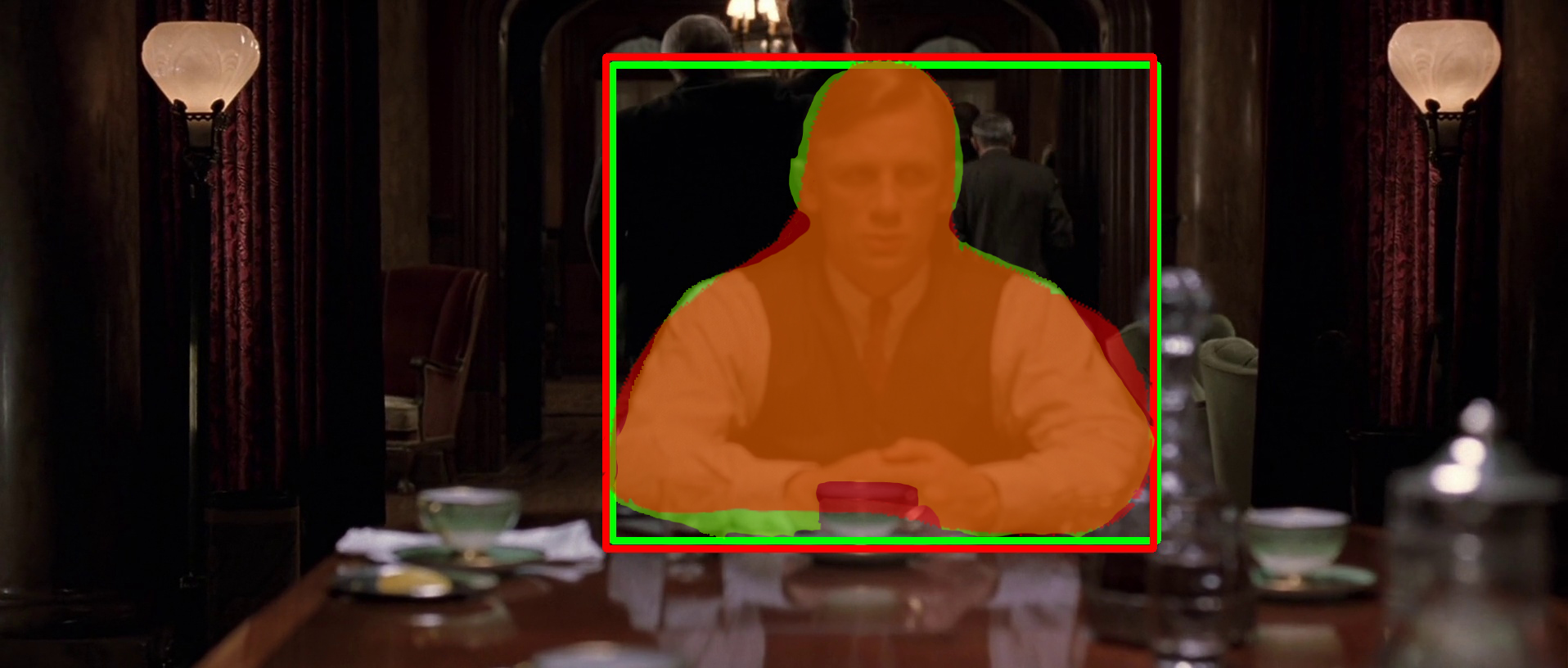} 

    \\
    
     \footnotesize (a)  & \footnotesize (b) 
     \\
     
     \includegraphics[width=0.40\columnwidth, height = 1.75cm]{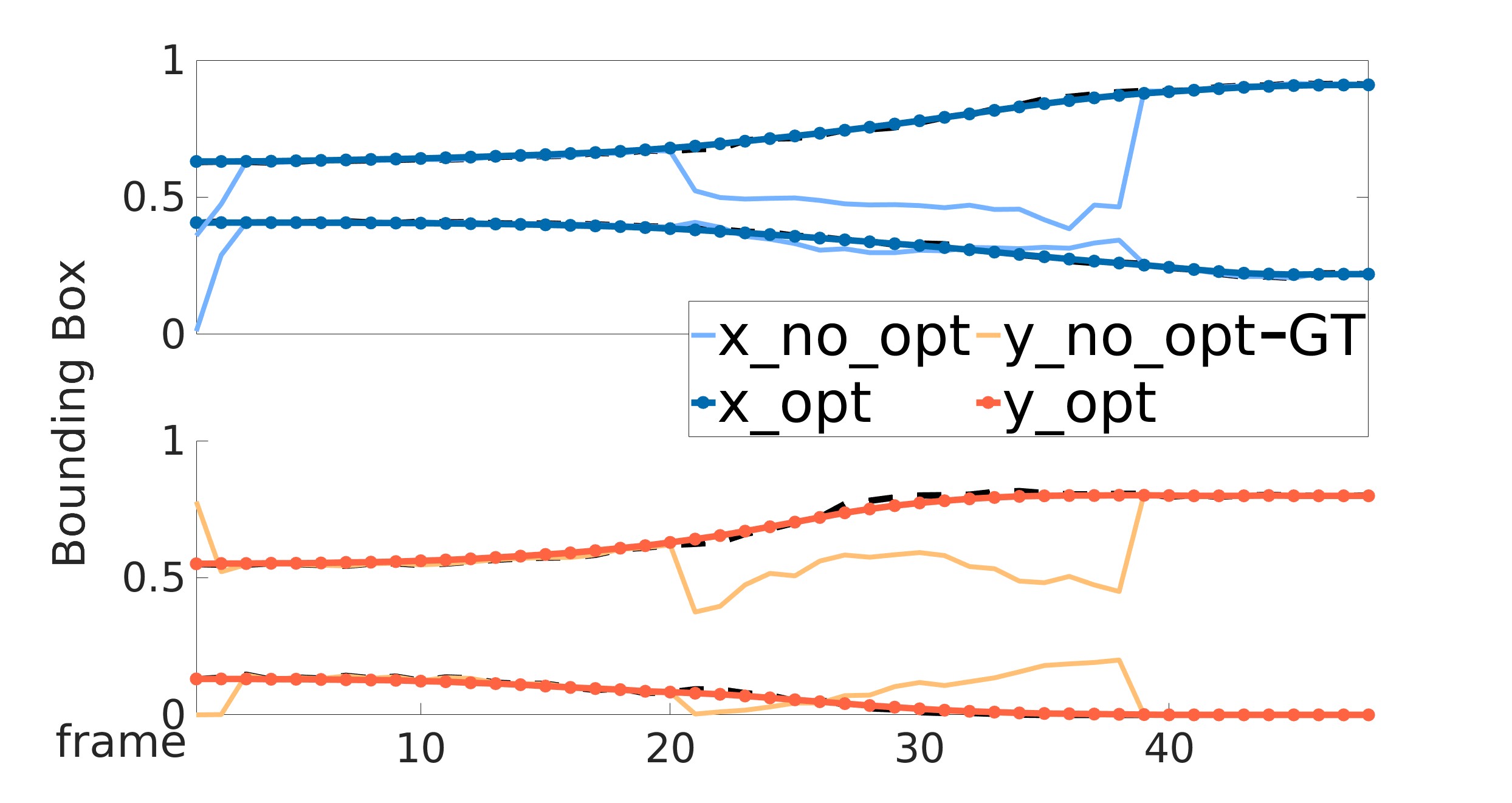}
    &
     \includegraphics[width=0.38\columnwidth]{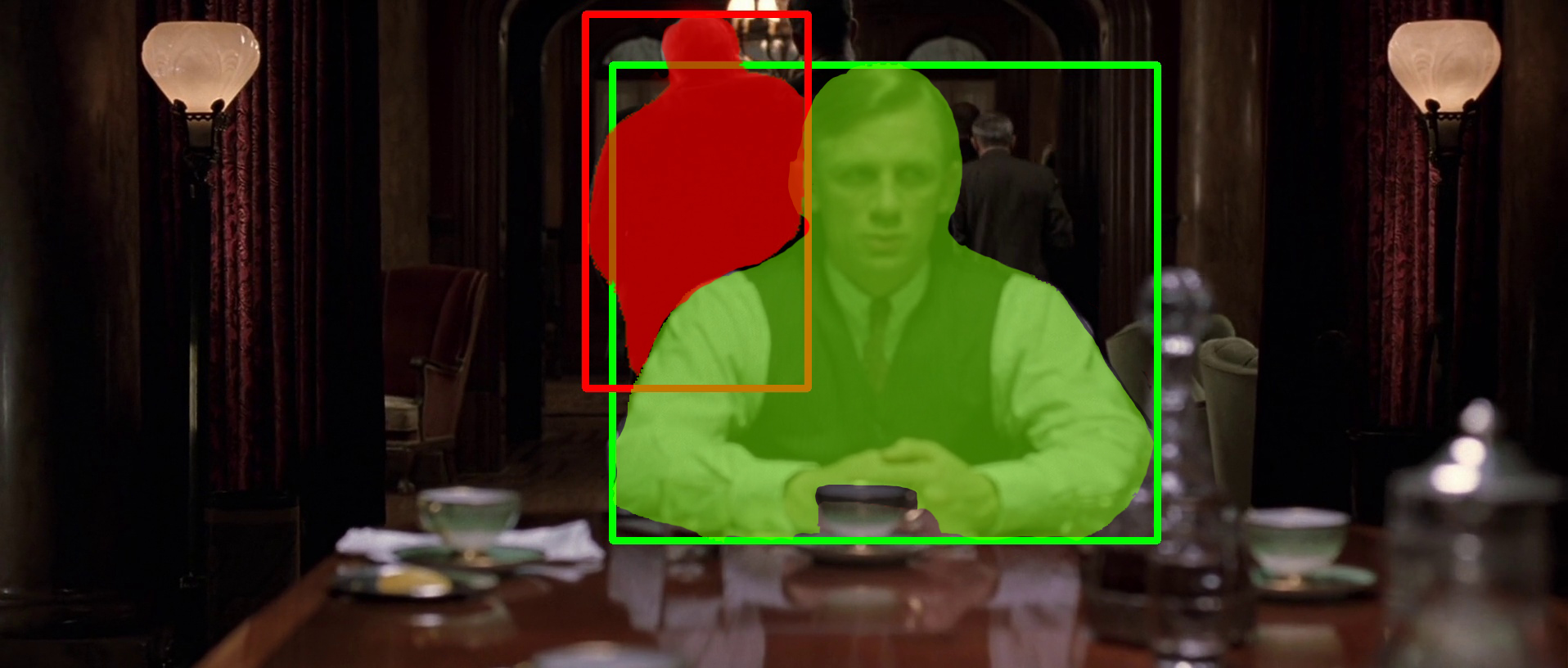}
    \\
     \footnotesize (c)  & \footnotesize (d) 
    
\end{tabular}
\caption{\textbf {Detection of main subject position and mask.} 
(a) Raw output from object segmentation library, i.e.,  all objects detected. (b) Optimized output of the subject localization module, with only one main subject selected. Green is ground truth and the red is optimized output. (d) Wrong selection of the main subject if we do not apply the optimization. 
Green is the ground truth and red is the output of our framework without the optimization. 
(c) Plot showing the components of the bounding box of the main subject. Blue lines are horizontal components and orange are vertical components. Dashed black lines are ground truth, light colors are the output not optimized, and  dark colors are the output optimized.} 
\label{fig:exp_image_position}
\end{figure}

We evaluate this feature quality on the real source sequence (S1) because it has 
manual labels of the bounding box and mask of the main subject. 
Figure~\ref{fig:exp_image_position}  shows the raw output of Mask-RCNN (3-a), the optimized detection (3-b) and the detection not optimized (3-d). In this case, the main subject appears blurry, so Mask-RCNN gives a low confidence, misleading the detection and selecting another subject. Figure~\ref{fig:exp_image_position}-c 
shows quantitative results along the complete sequence. 
 
%

%
%

\subsubsection{\textbf{Detection of body joints}}
\begin{figure}[!bt]
\centering
\begin{tabular}{cc}
    \includegraphics[width=0.35\columnwidth]{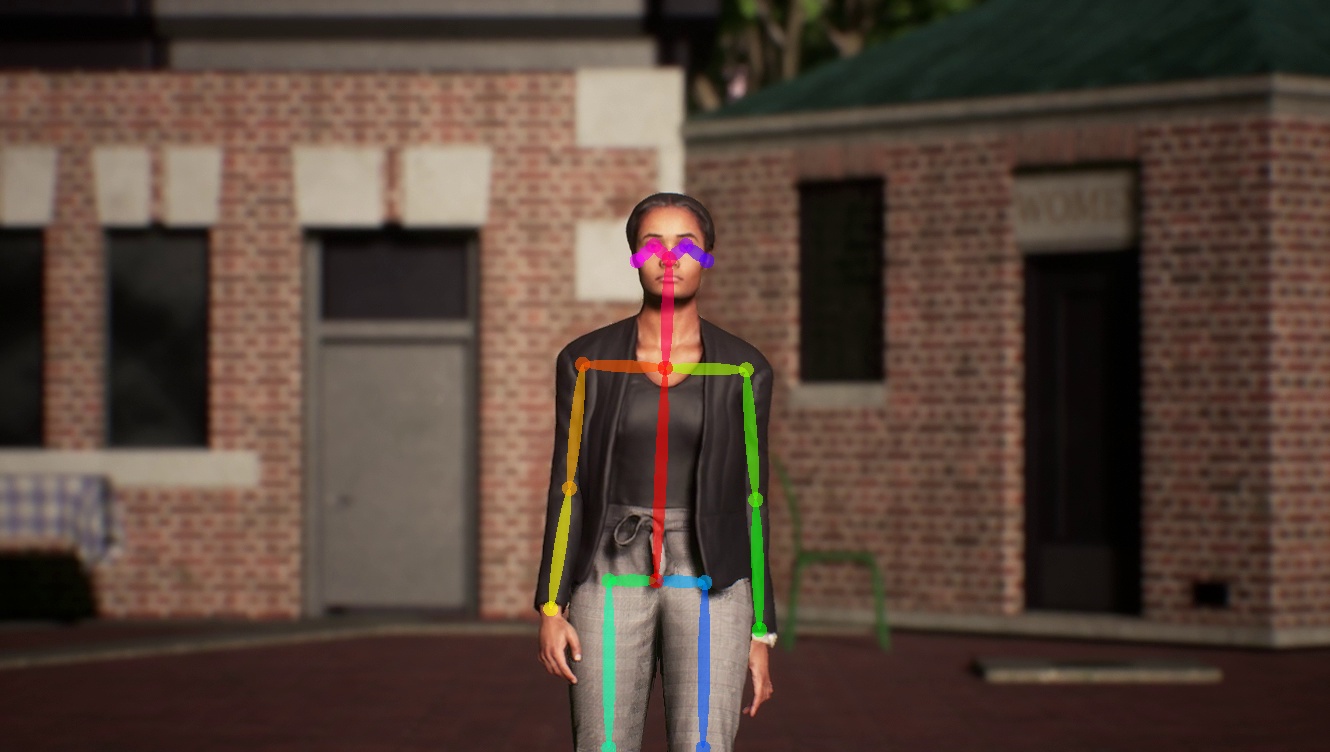}
    &
    
     
    \includegraphics[width=0.4\columnwidth]{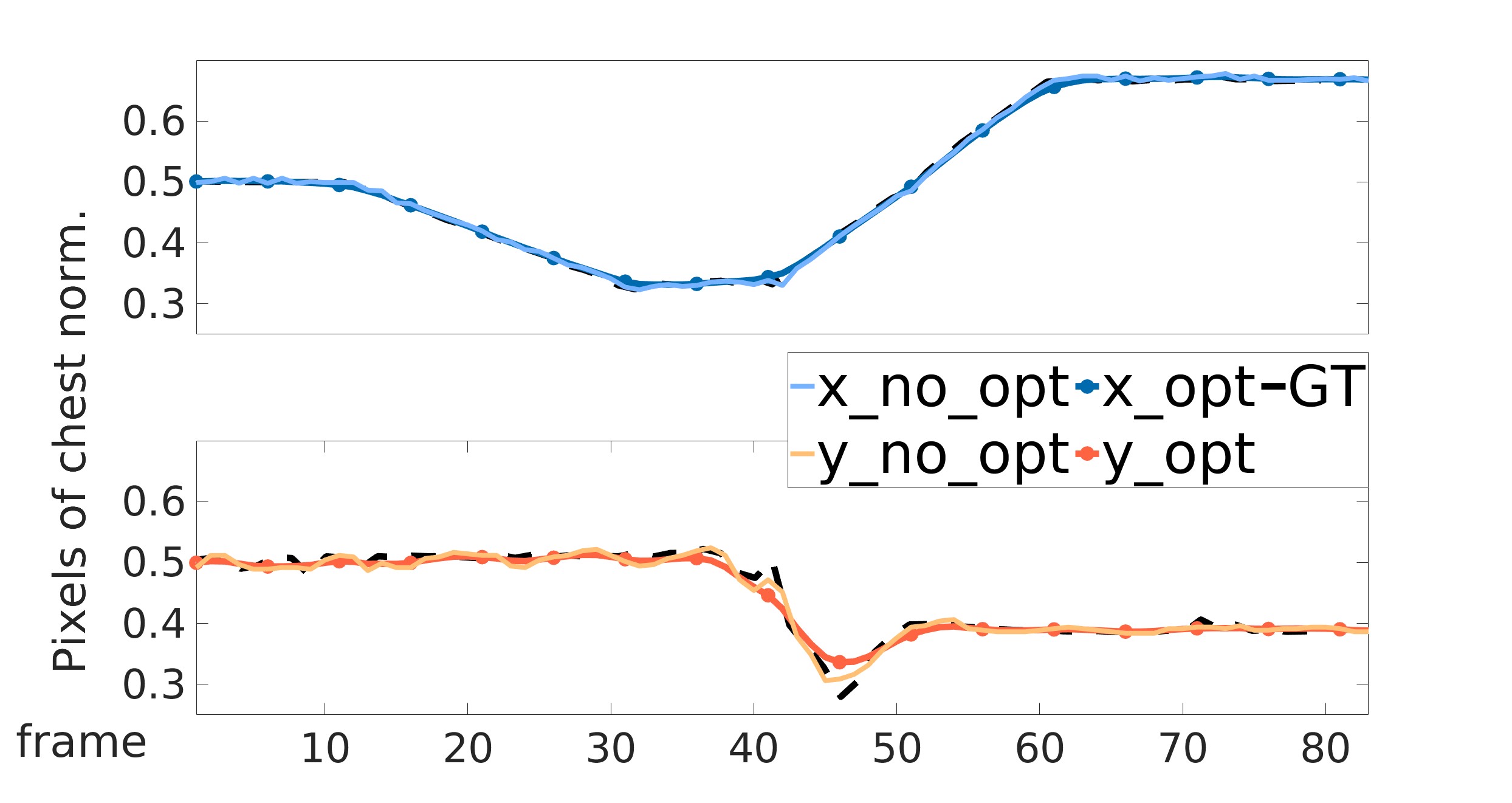}
    \\ 
     \footnotesize (a)  & \footnotesize (b) \\
\end{tabular}
\centering\\
\caption{\textbf {Detection of body joints.} (a) Skeleton with the body joints for one frame (b) Plot showing the two components for the chest of the subject (red middle marker in skeleton). Blue lines are horizontal components and orange are vertical components. Dashed black lines are ground truth, light colors are the output not optimized (with noise), and  dark colors are the output optimized. }
\label{fig:body_joints}
\end{figure}

 We validate the detection and optimization of the position of the body joints of the main subject in S2 because we have accurate ground truth information.
 Figure \ref{fig:body_joints}-a shows a frame together with a graphical representation of the body joints.
 Figure \ref{fig:body_joints}-b depicts the evolution of the position on the image of the person's chest, 
 which is a representative body joint of a person. 
 The optimization algorithm removes the noise from the measurement, producing a stable measure close to the ground truth. 
 Table \ref{tab:table_body_joints1} shows the error between the output of the framework and the ground truth values when the output is optimized (first row) and not optimized (second row). 

\begin{table}[!b]
    \centering
    \begin{tabular}{|c|c|c|c|c|}
    \hline
&mn $x$ & std $x$ &  mn $y$  & std $y$  \\
    \hline
S2.Opt & 0.065 & 0.053 & 0.055 & 0.110  \\
    \hline 
S2.Raw & 0.066 & 0.054 & 0.056 & 0.111   \\
\hline
    \end{tabular}
    \caption{Mean and standard deviation of the error (in pixels) in chest body-joint from main subject}
    \label{tab:table_body_joints1}
\end{table}

\begin{table}[!hb]
    \centering
    \begin{tabular}{|c|c|c|c|c|c|c|}
    \hline
&mn \textit{bg} & std \textit{bg} &  mn \textit{ms}  & std \textit{ms}  &  mn \textit{fg}  & std \textit{fg}\\
    \hline
S1.Opt & 0.057 & 0.117 & 0.061 & 0.115 & 0.001 & 0.001   \\
    \hline 
S1.Raw & 0.275 & 0.113 & 0.107 & 0.141 & 0.040 & 0.050   \\    \hline
S2.Opt & 0.027 & 0.097 & 0.039 & 0.146 & 0.001 & 0.001   \\
    \hline 
S2.Raw & 0.180 & 0.137 & 0.143 & 0.147 & 0.031 & 0.041   \\
\hline
    \end{tabular}
    \caption{Mean and standard deviation of error in percentage of focused pixels of the background (\textit{bg}), main subject (\textit{ms}), and foreground (\textit{fg}) for source-1 (S1) and source-2 (S2) seqs. }
    \label{tab:focus1}
\end{table}  
\subsubsection{\textbf{Focus of the scene}}
\begin{figure}[!ht]
\centering
\begin{tabular}{ccc}
 \includegraphics[width=0.30\columnwidth]{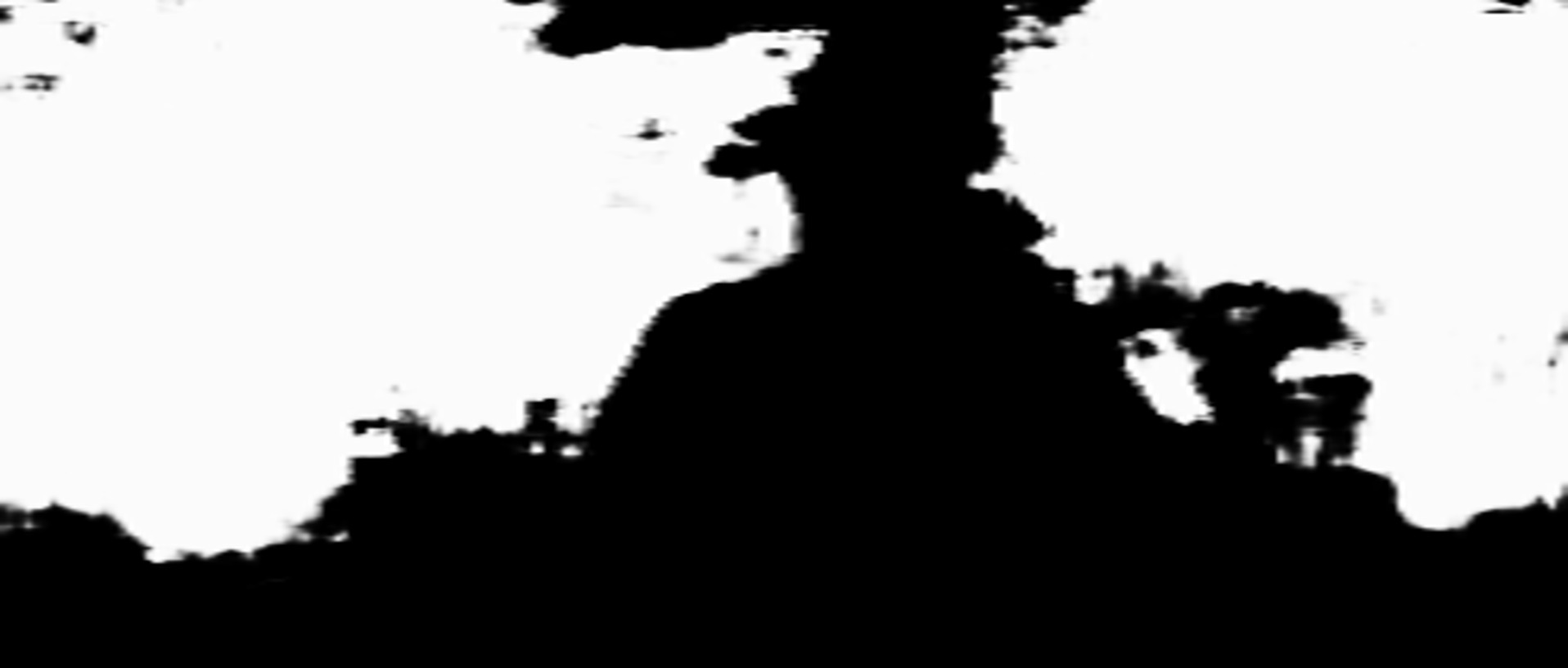} 
    &
    
    \includegraphics[width=0.30\columnwidth]{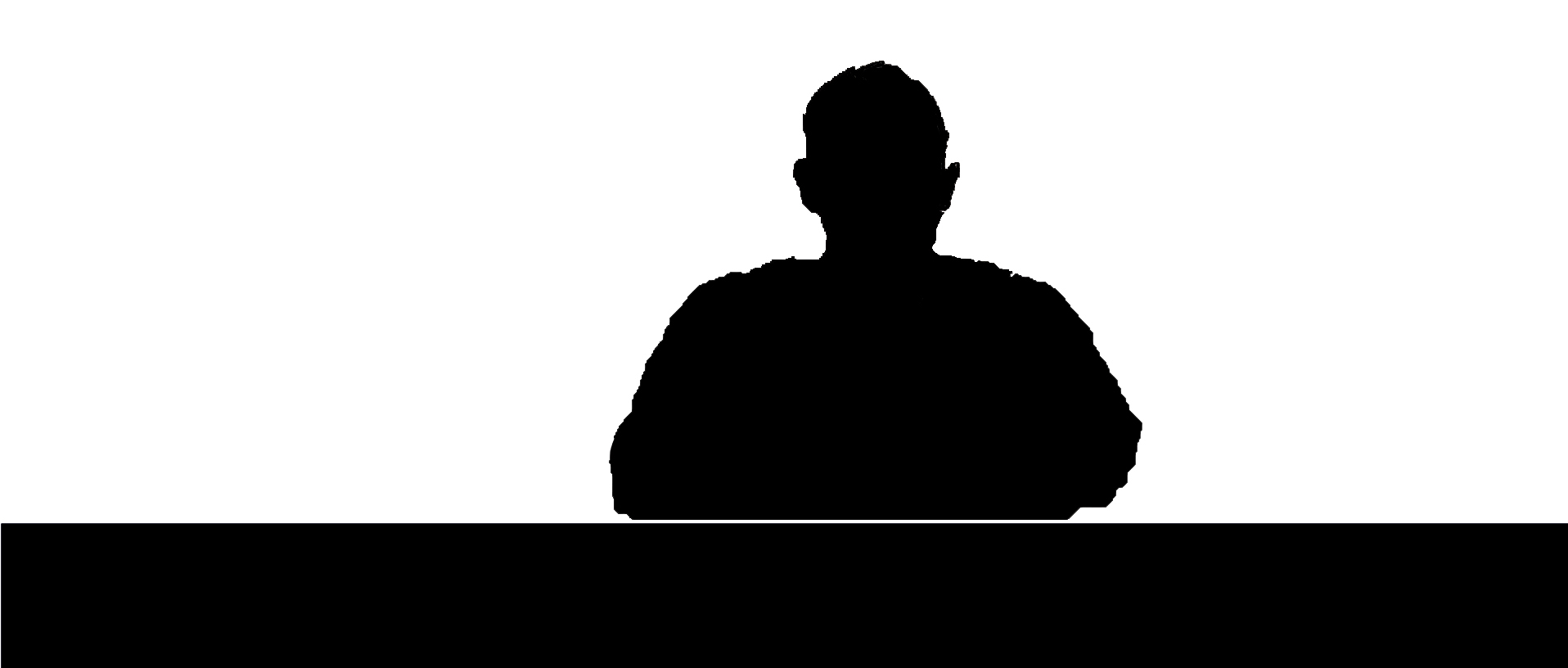} 
    &
    
    \includegraphics[width=0.30\columnwidth]{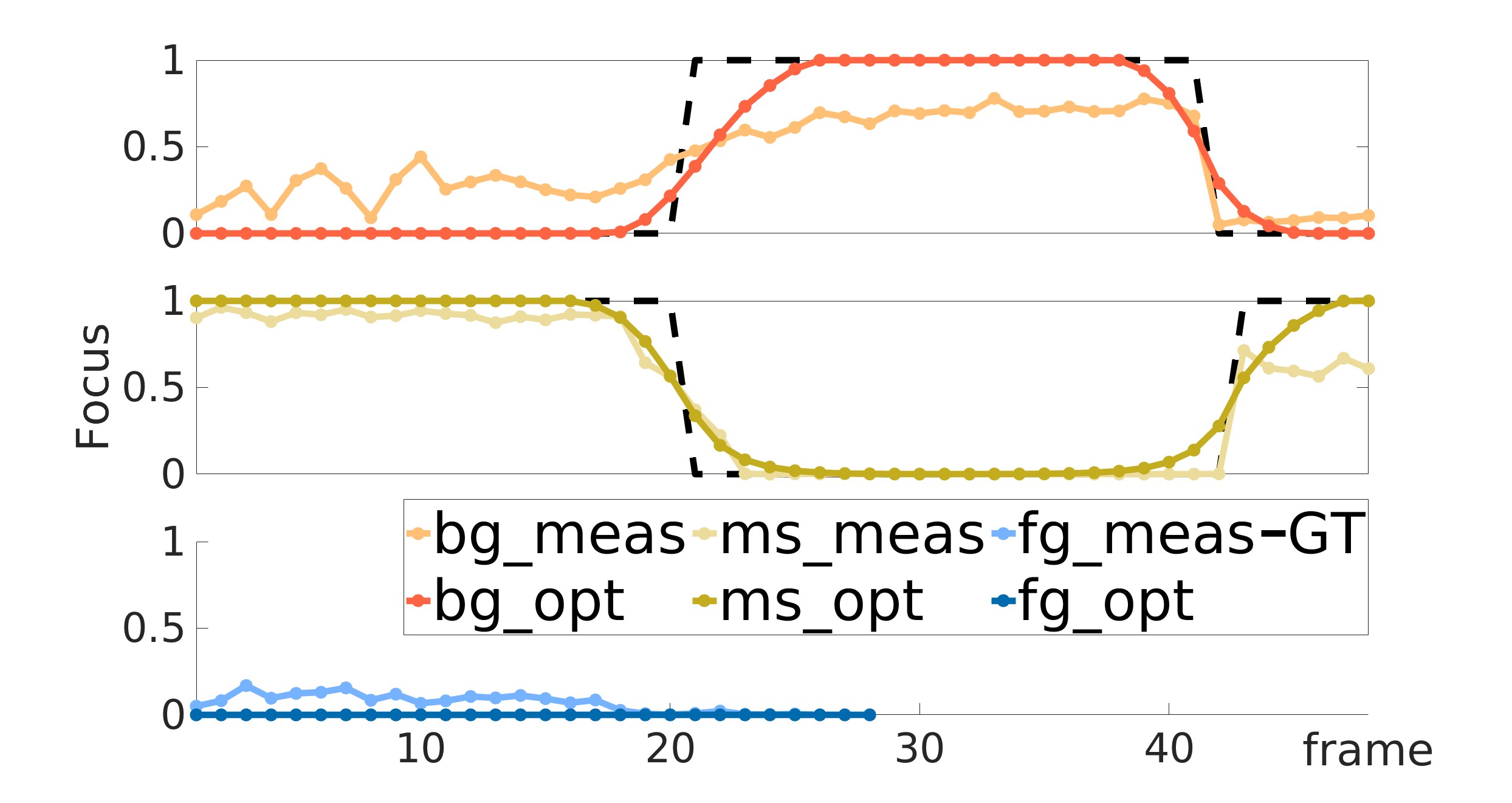} 
    \\
    
  \includegraphics[width=0.3\columnwidth]{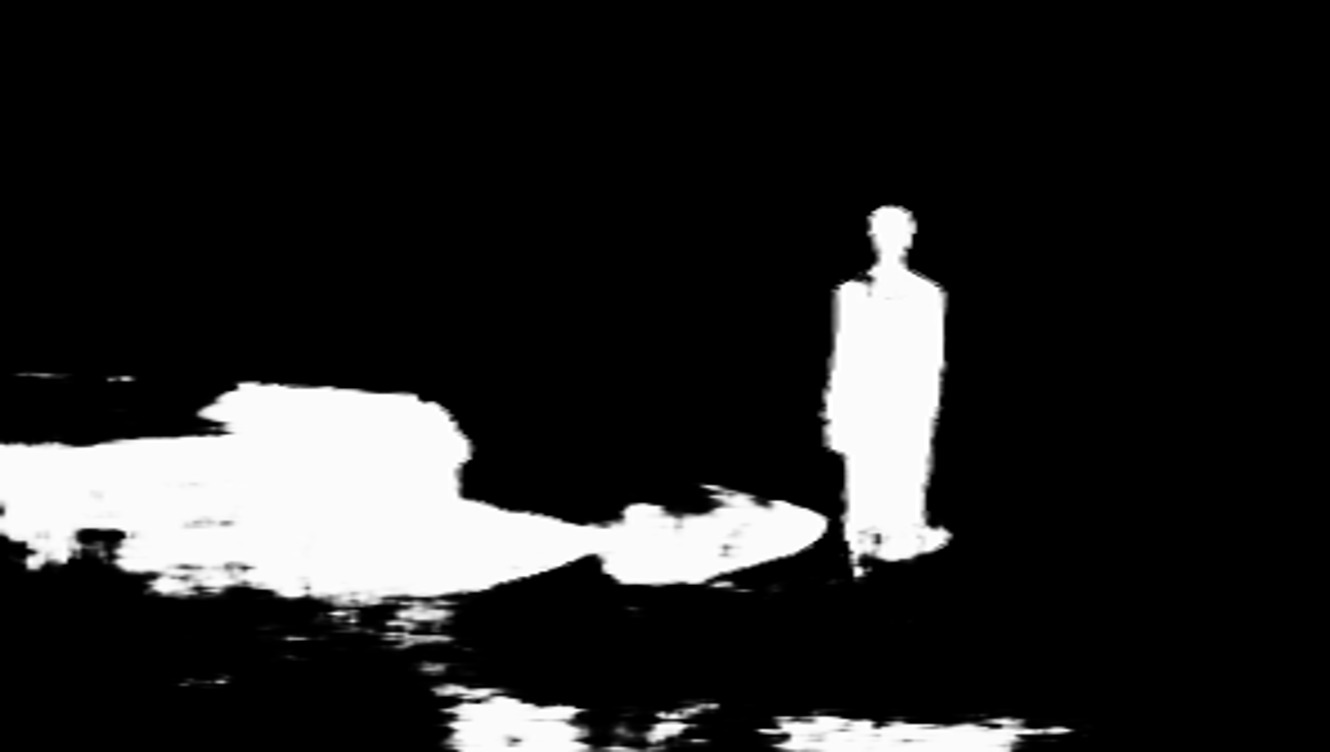}
    &
    \includegraphics[width=0.3\columnwidth]{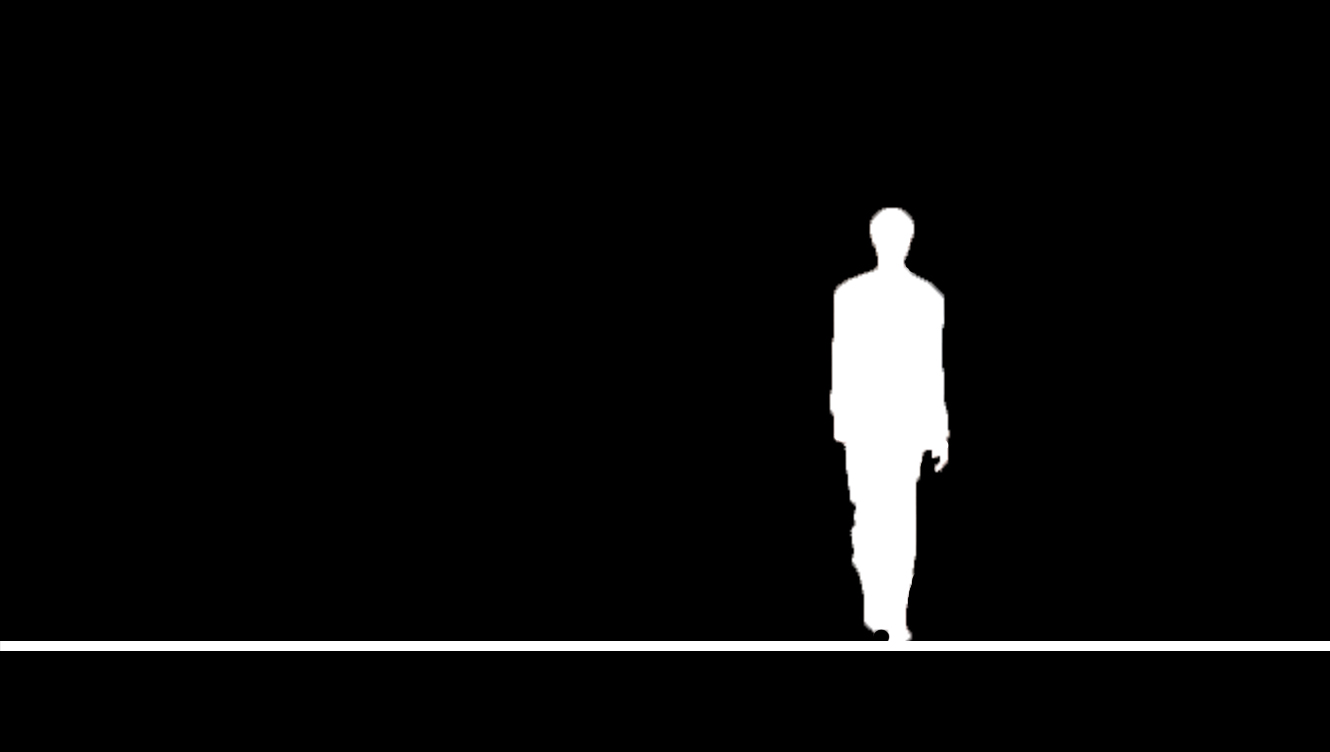}
    &
     \includegraphics[width=0.3\columnwidth]{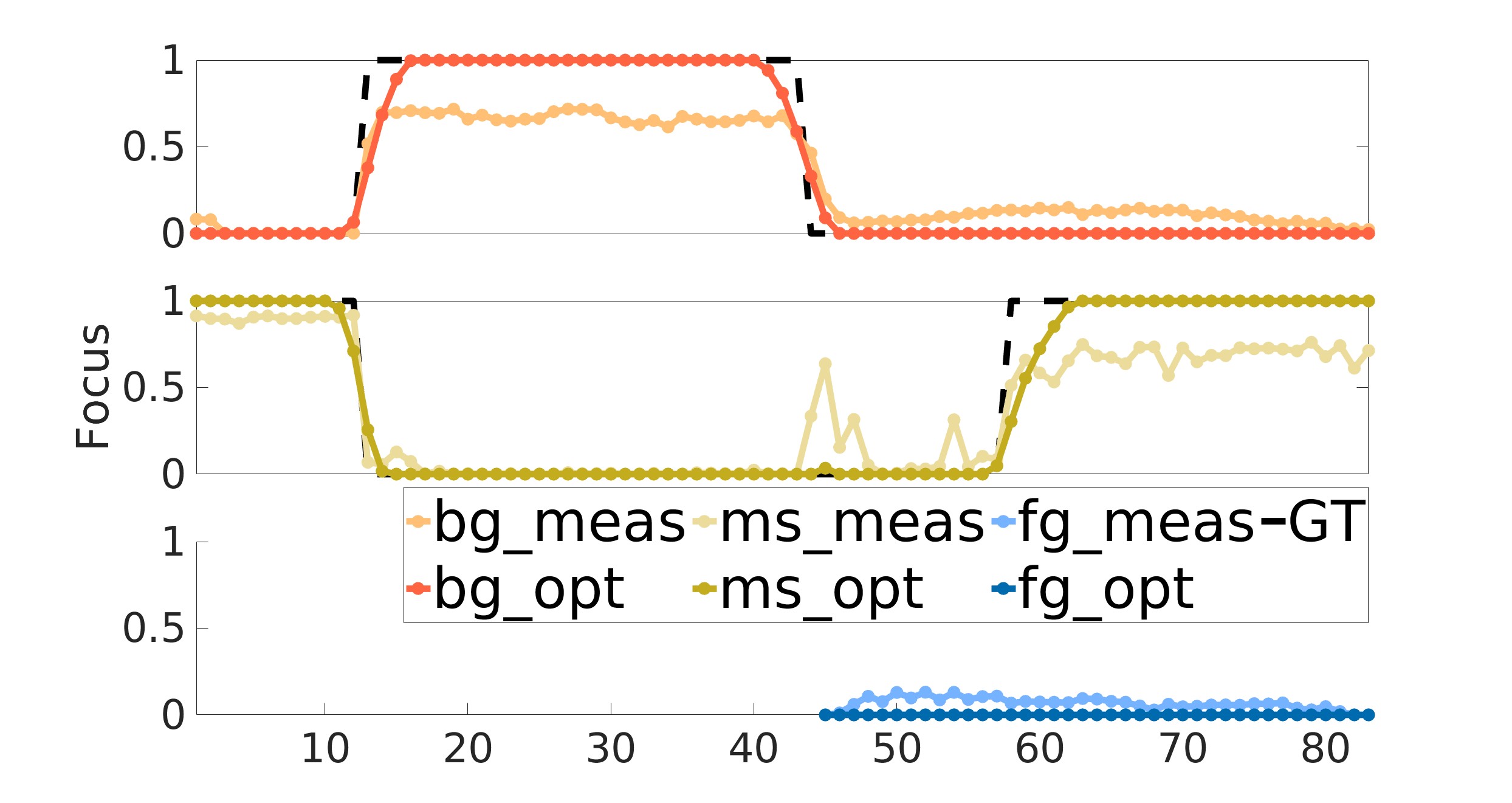}
    \\
     \footnotesize (a)  & \footnotesize (b) & \footnotesize (c) \\
     
\end{tabular}
\caption{\textbf {Focus of the scene.} 
First row is for S1 and second row is for S2. (a) Blur detection for a sample frame of each sequence. Black depicts pixels out of focus and white depicts pixels in focus. 
(b) Optimized output of the focus detection module. 
(c) Evolution of the ground truth (black dashed line), the raw measure (light solid lines), and the optimized output (solid dark lines) of the percentage of focused pixels. Orange is the background, yellow is the subject, and blue is the foreground (only present when full body of subject is shown), '1' is focused and '0' is not focused.}
\label{fig:focus_analysis}
\end{figure}

To evaluate the detection of the depth of field, we manually labeled S1, annotating if the foreground, subject, and background are in focus or not. This information is also available for S2, since it is simulated. 
For both sequences, we compare the real labels with the output of our framework. 
Figure \ref{fig:focus_analysis} shows  qualitative and quantitative results. 
 Table \ref{tab:focus1} shows the error between the output of the focus given by the framework and their ground truth values. 

\subsection{Style Transfer}
\begin{figure}[!ht]
\centering
\begin{tabular}{ccc}
     \includegraphics[width=0.27\linewidth]{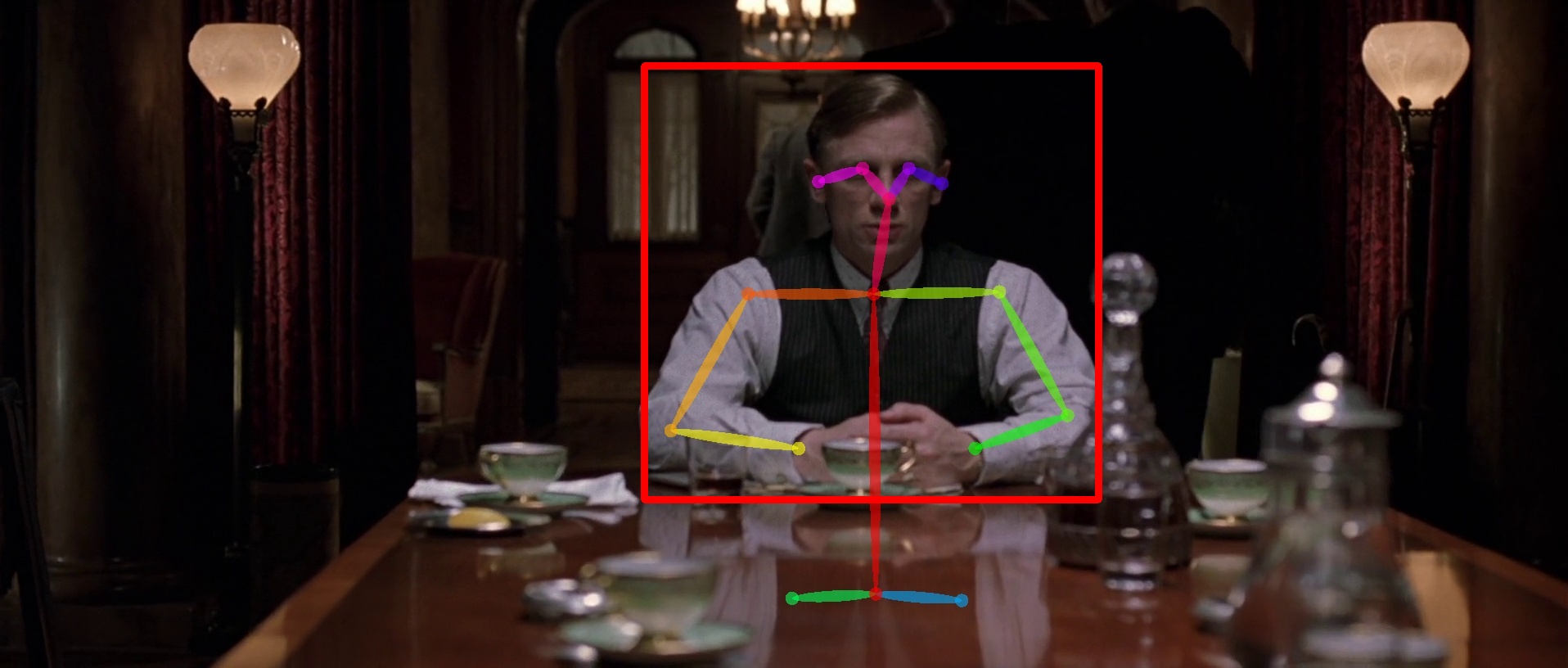}
    & \includegraphics[width=0.27\linewidth]{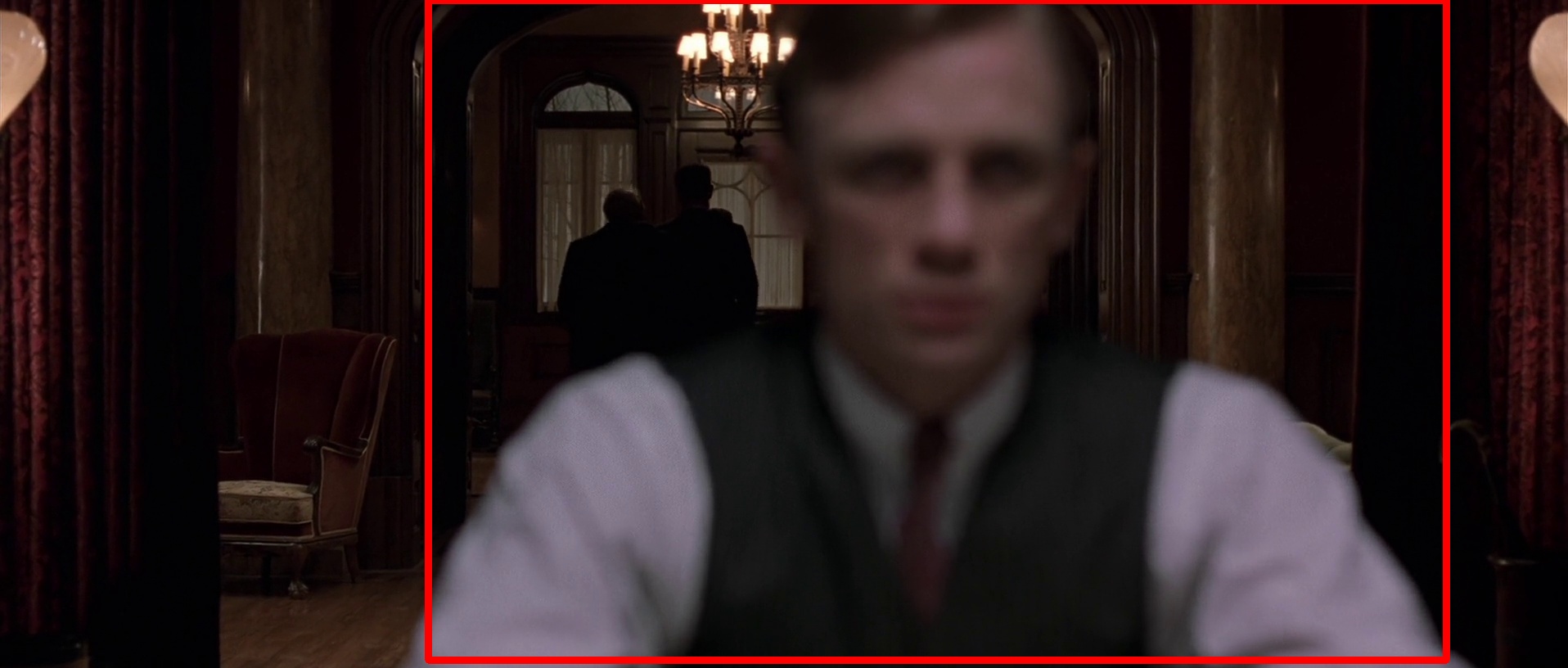}
    & \includegraphics[width=0.27\linewidth]{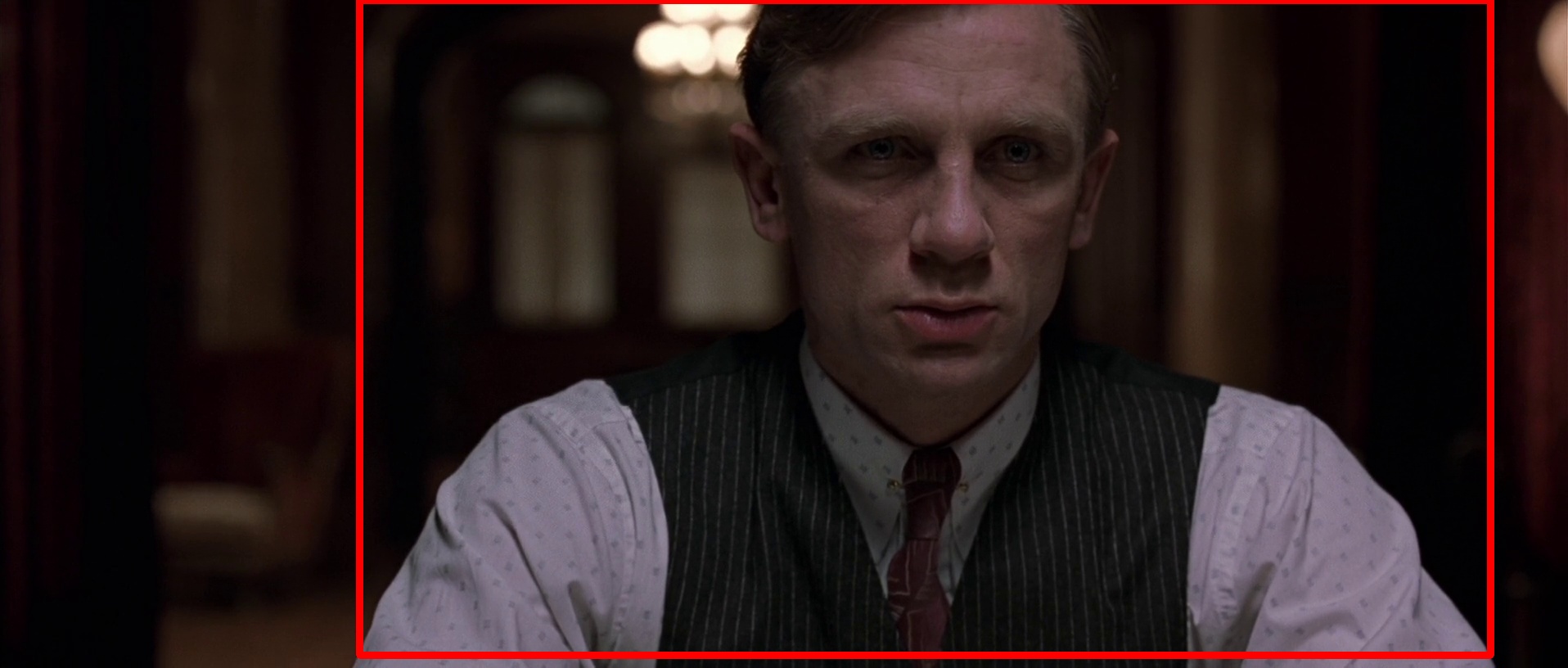}
    \\
     \includegraphics[width=0.27\linewidth]{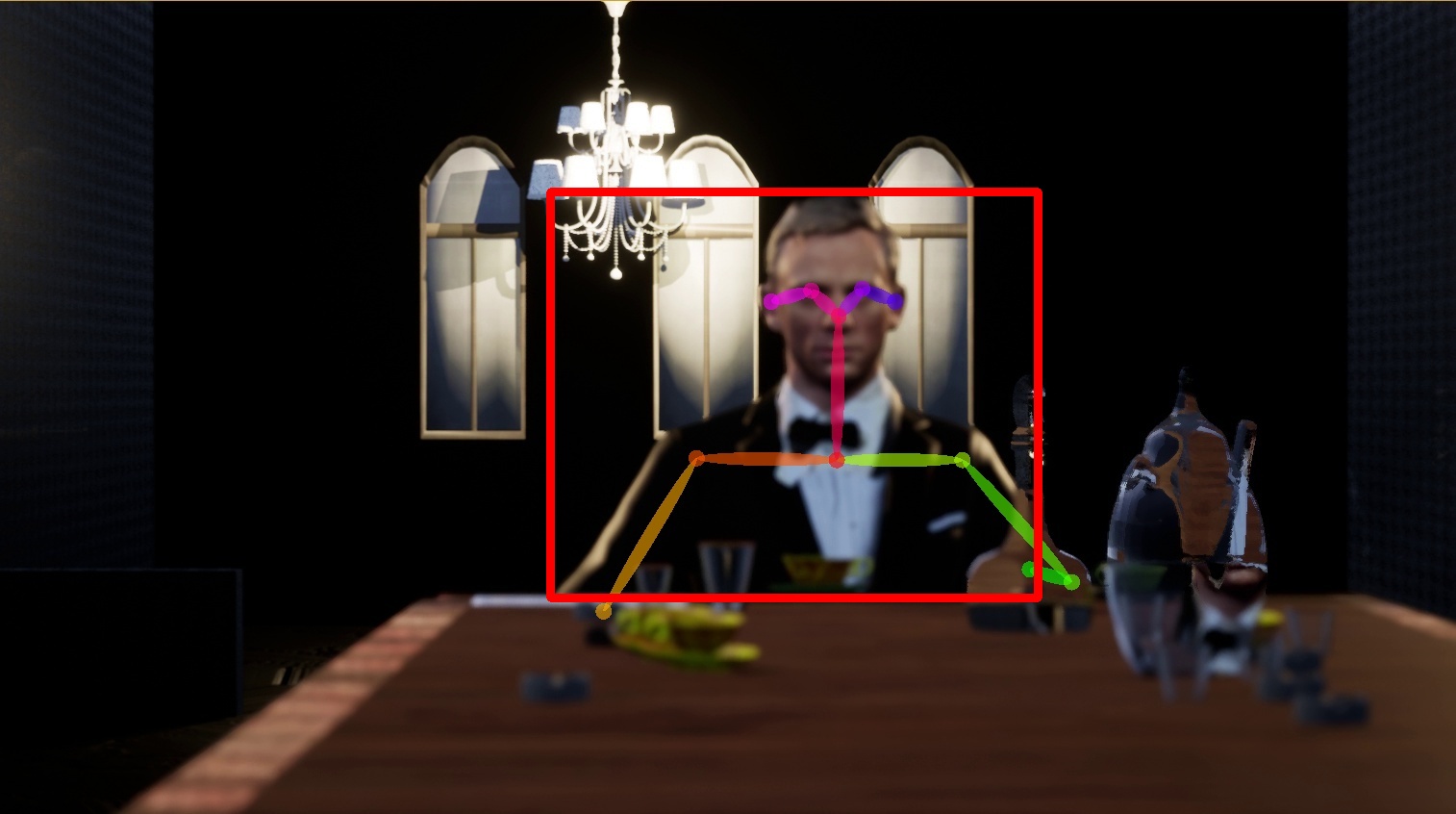}
    & \includegraphics[width=0.27\linewidth]{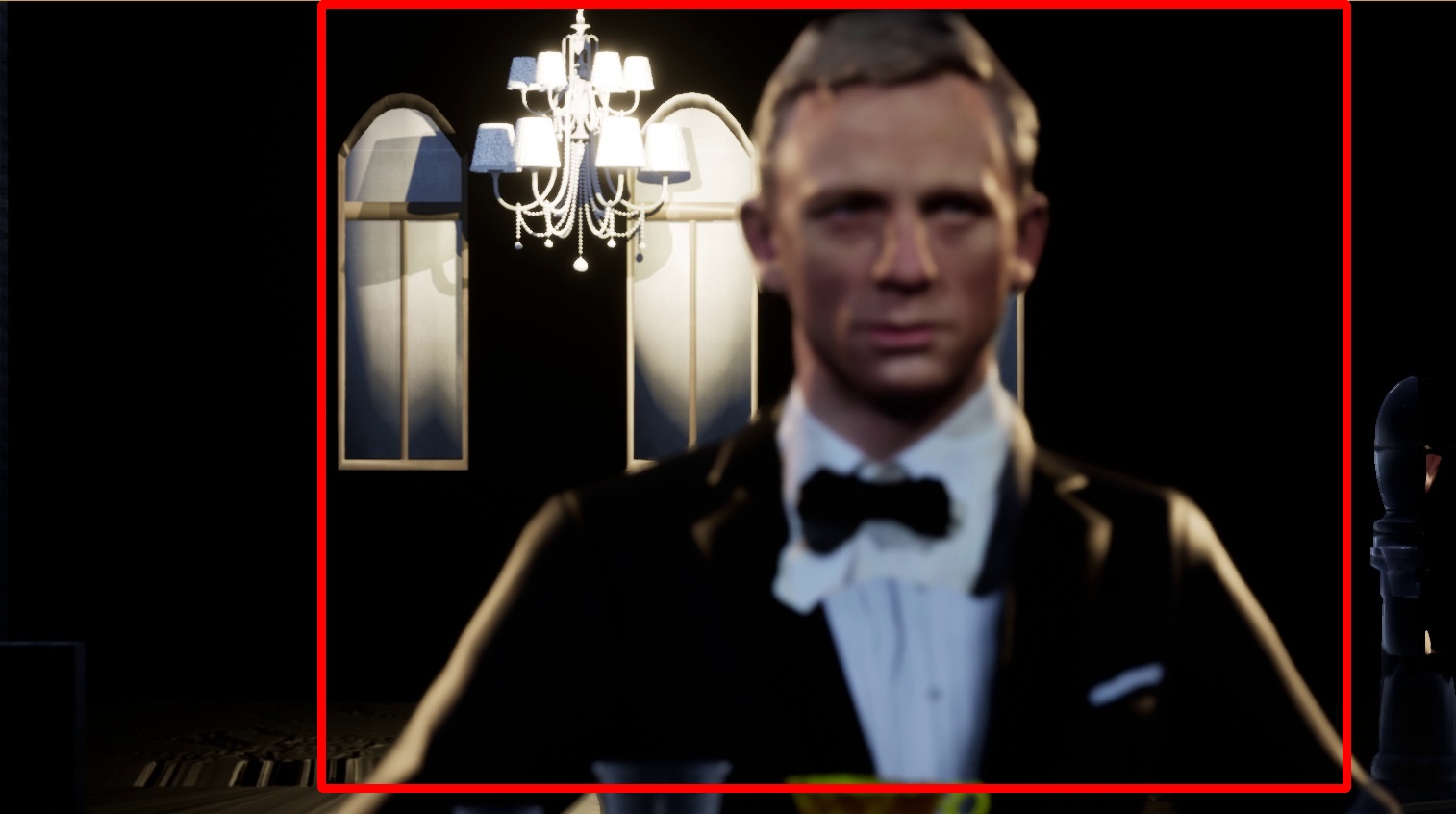}
    & \includegraphics[width=0.27\linewidth]{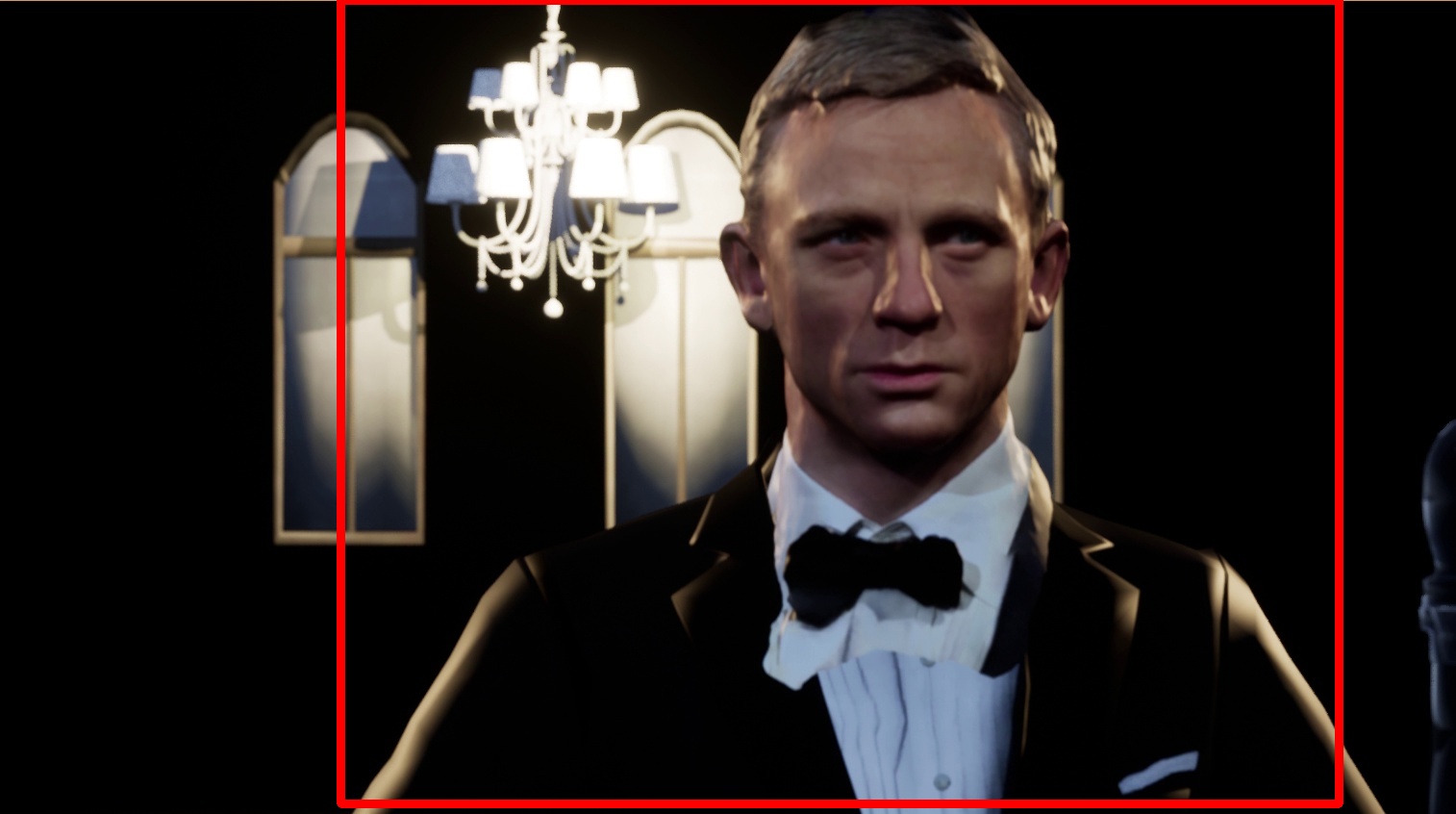}
    \\
     \includegraphics[width=0.27\linewidth]{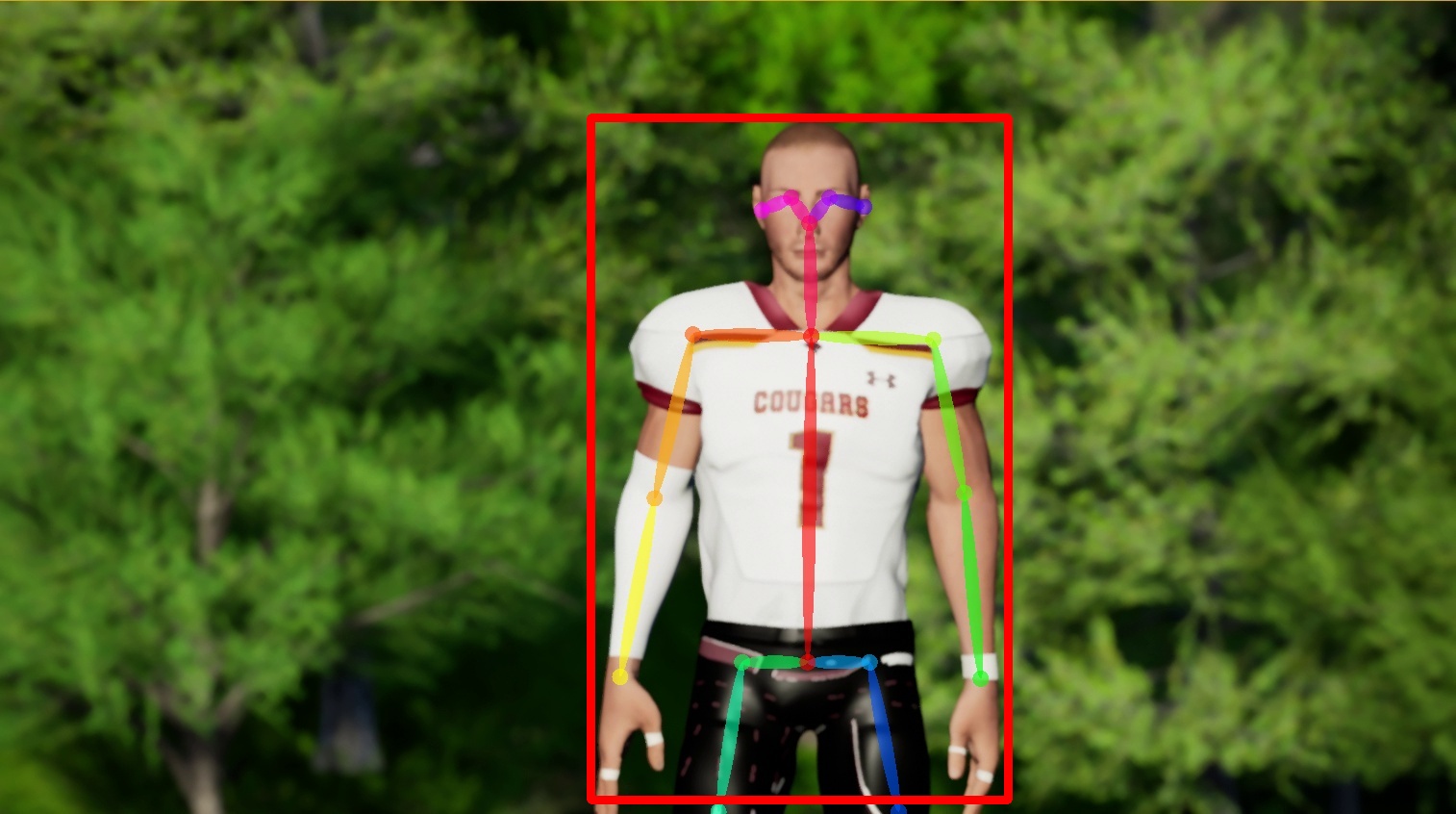}
    & \includegraphics[width=0.27\linewidth]{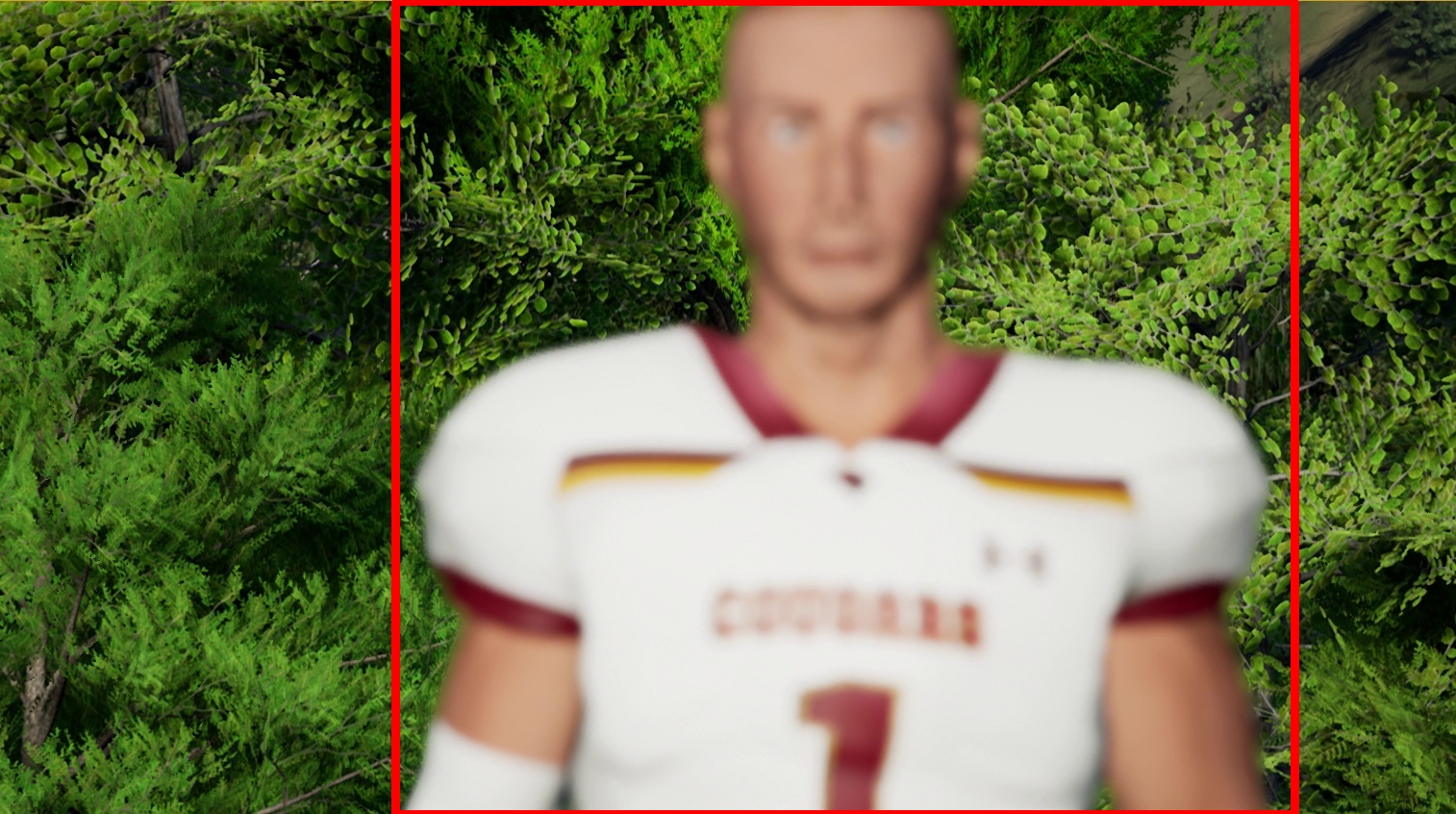}
    & \includegraphics[width=0.27\linewidth]{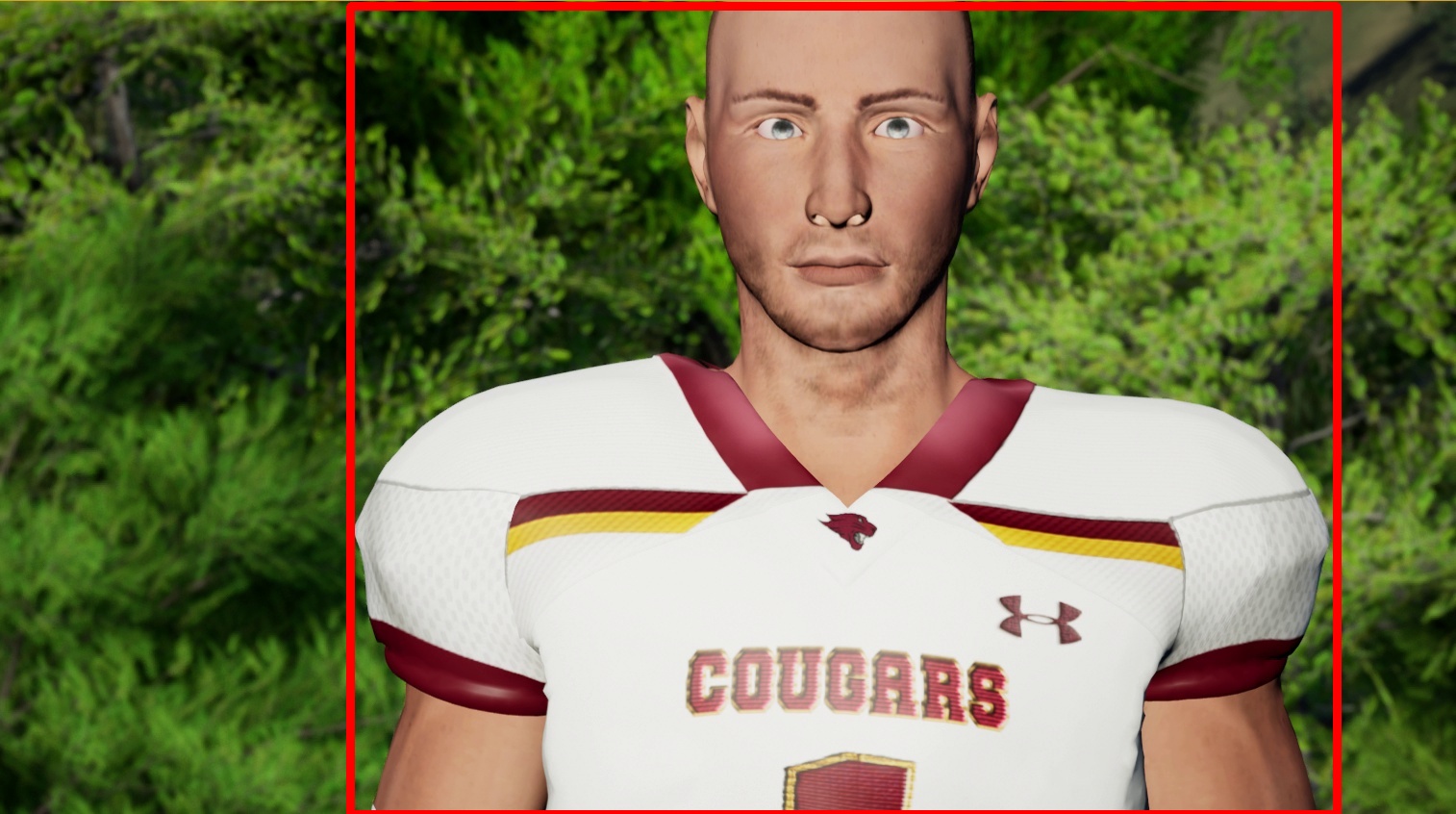}
\end{tabular}
\caption{\textbf {Style Transfer from S1 to T1 and T3.}   First row: sampled frames 
from the source ordered in time left to right. Second and third row: corresponding frames from the obtained output sequences in T1 and T3.} 
\label{fig:exp_comparing_real}
\end{figure}

\begin{figure}[!ht]
\centering
\begin{tabular}{ccccc}
    \includegraphics[width=0.27\linewidth]{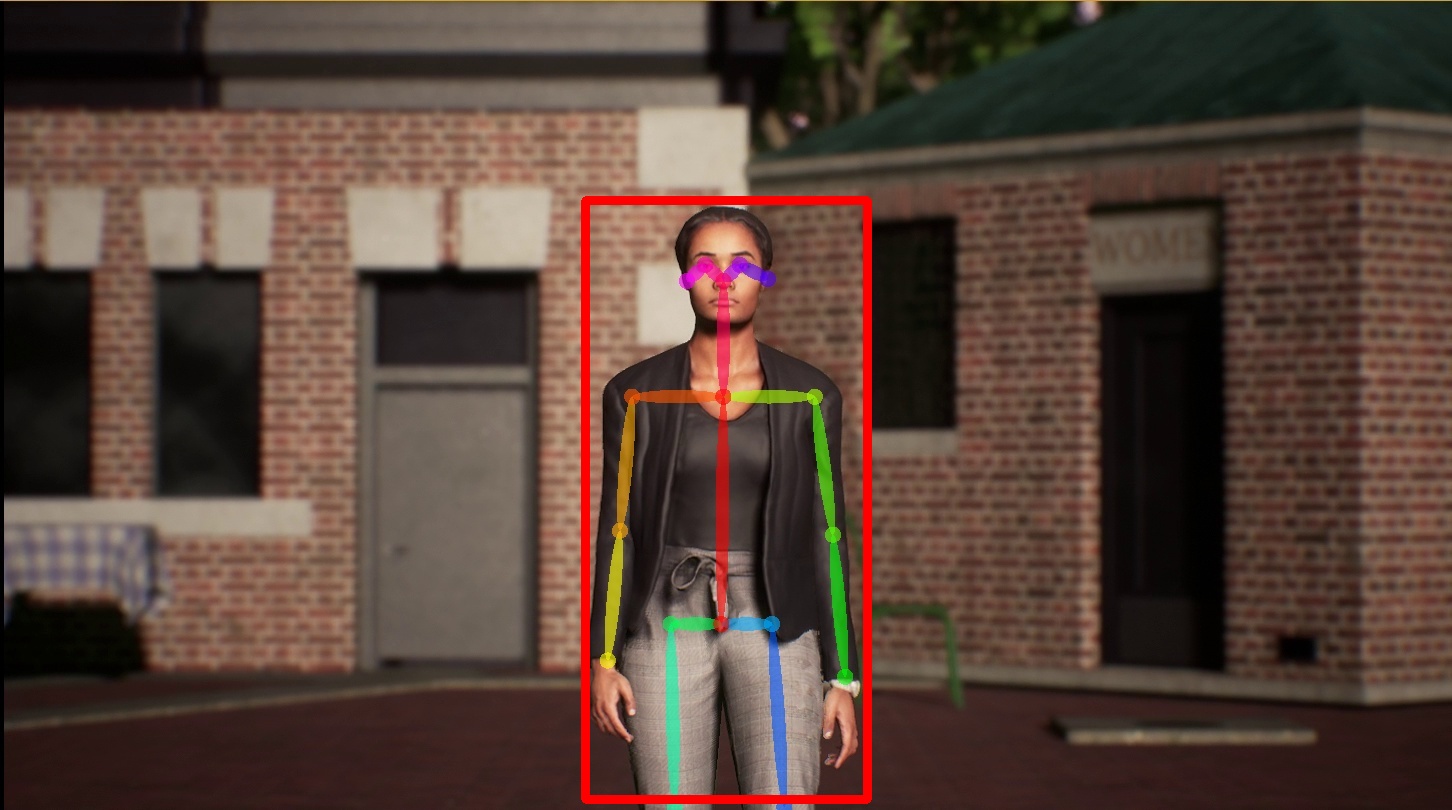}
    & \includegraphics[width=0.27\linewidth]{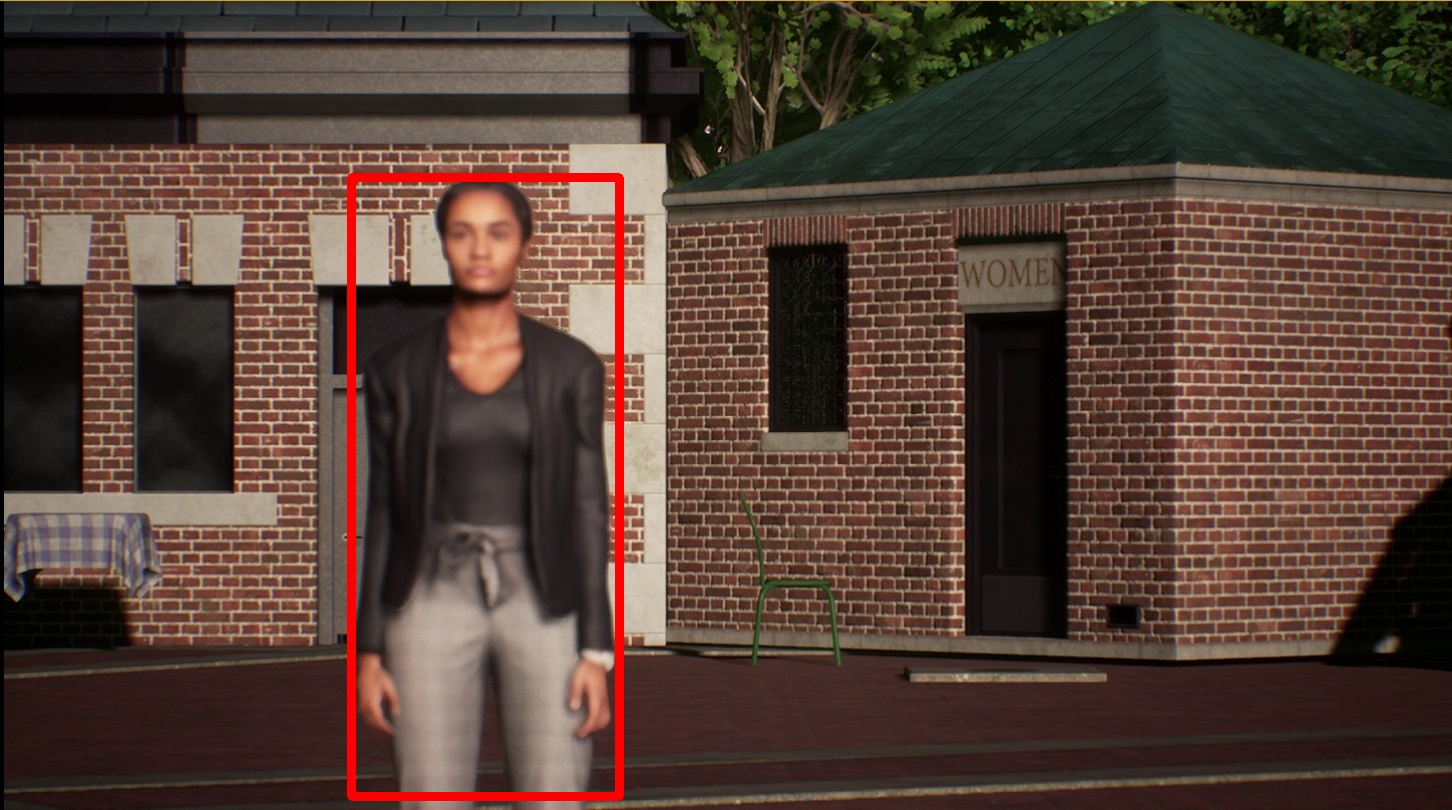}
    & \includegraphics[width=0.27\linewidth]{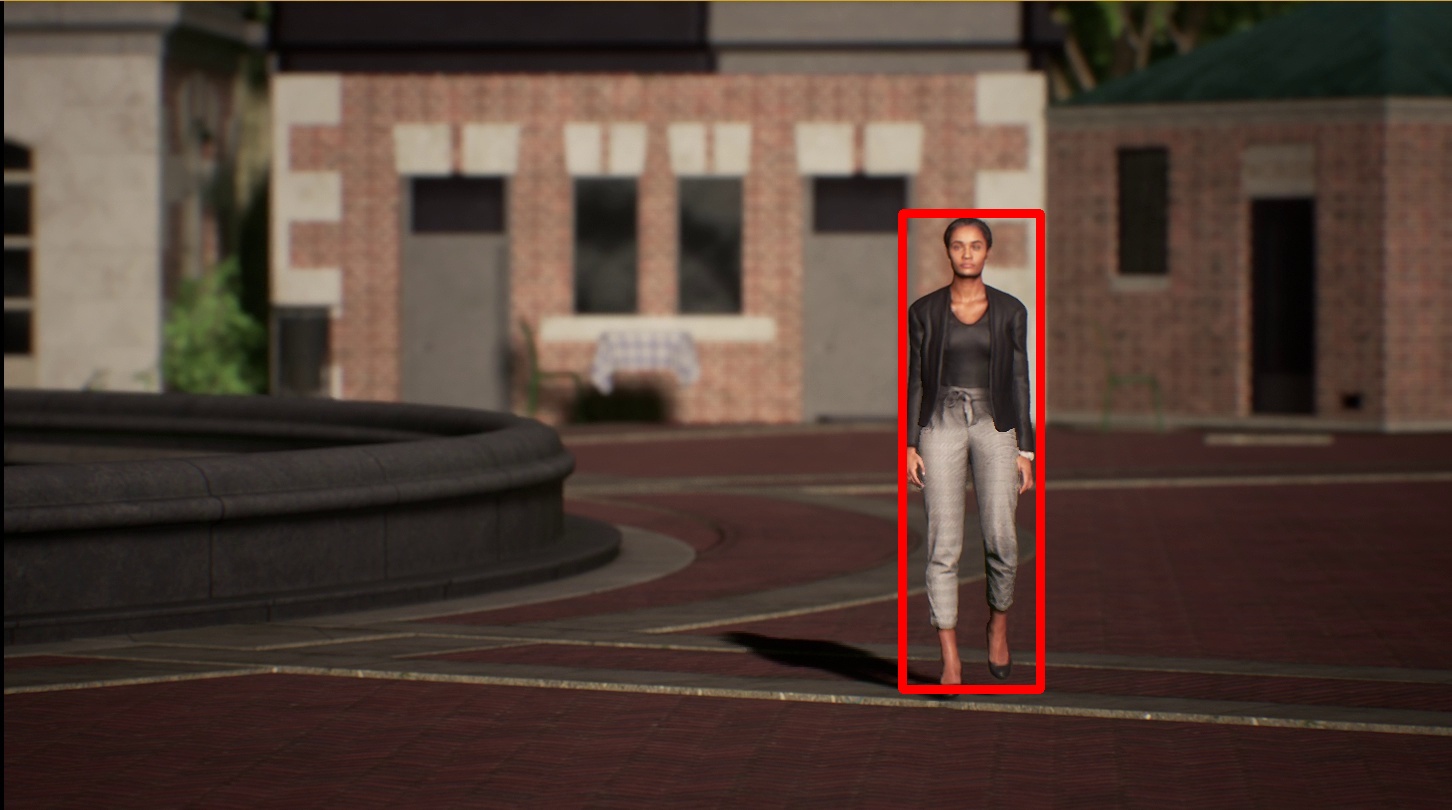}
    \\
     \includegraphics[width=0.27\linewidth]{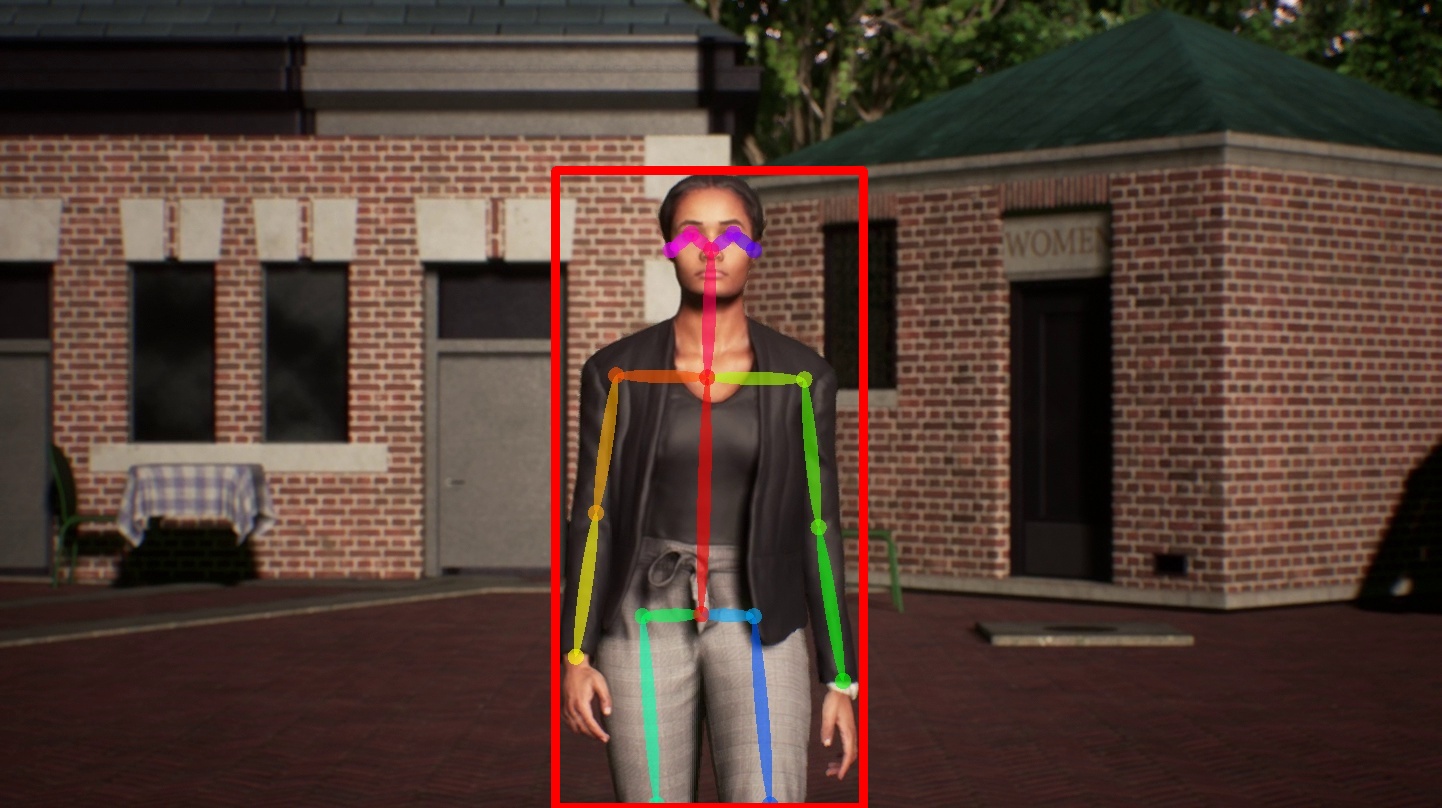}
    & \includegraphics[width=0.27\linewidth]{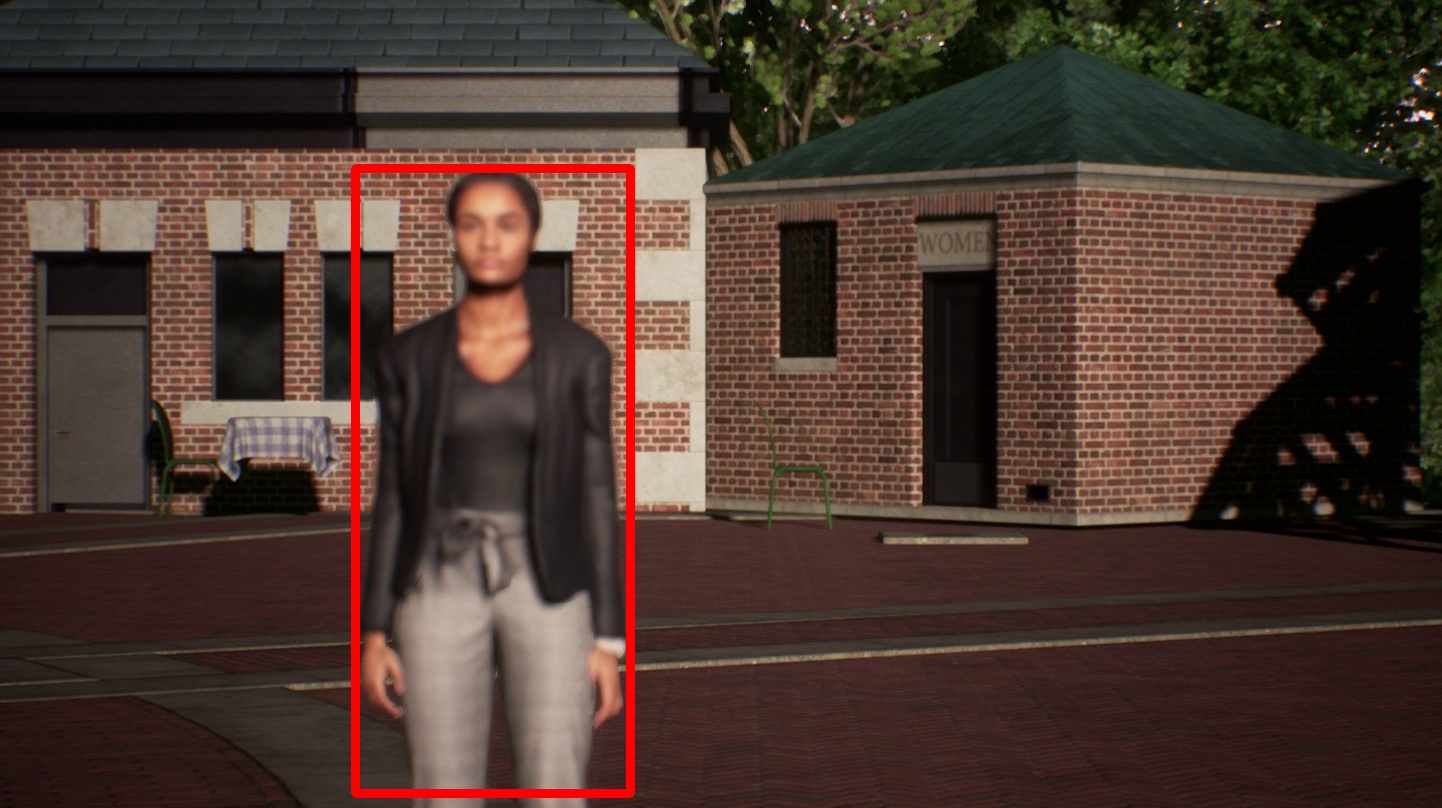}
    & \includegraphics[width=0.27\linewidth]{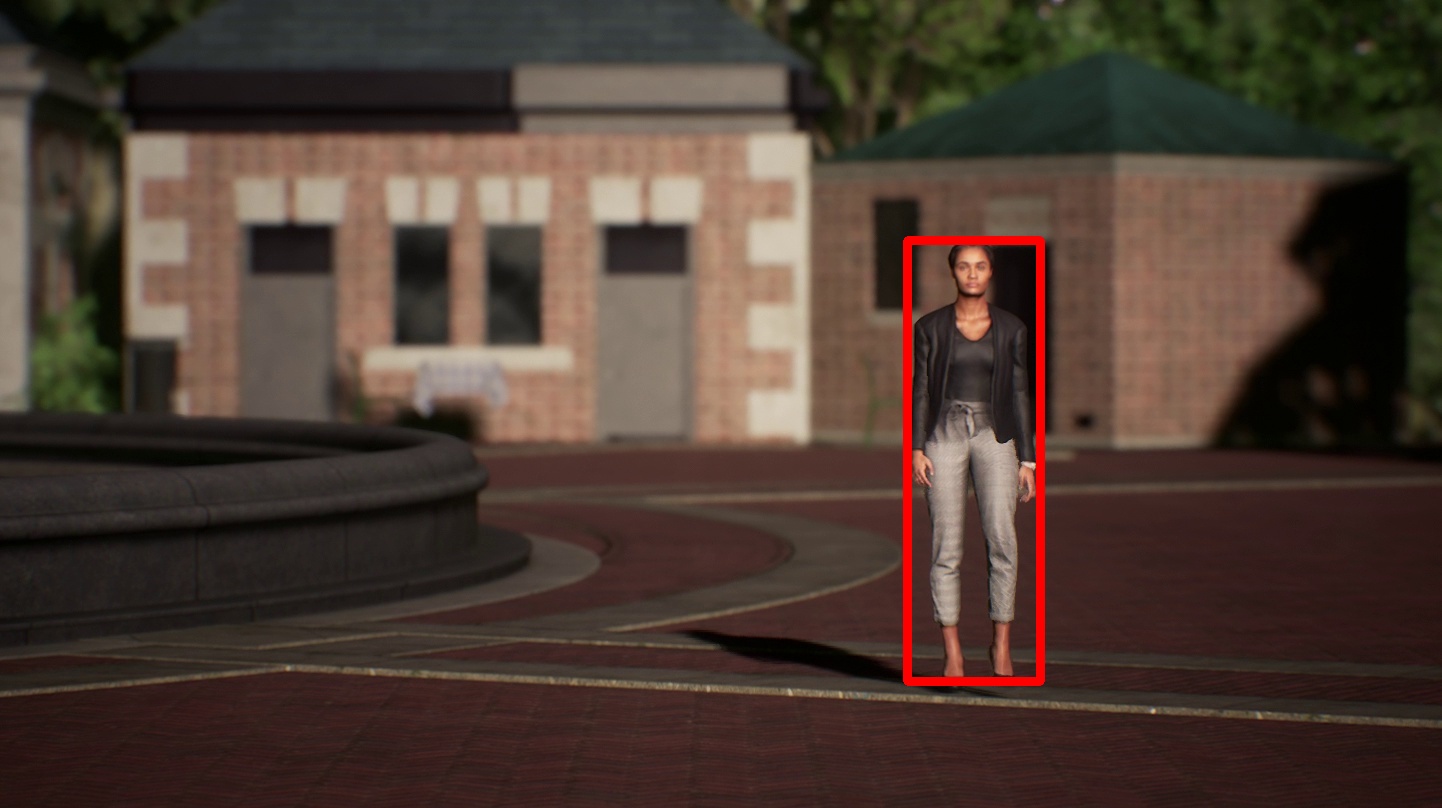}
    
    \\
     \includegraphics[width=0.27\linewidth]{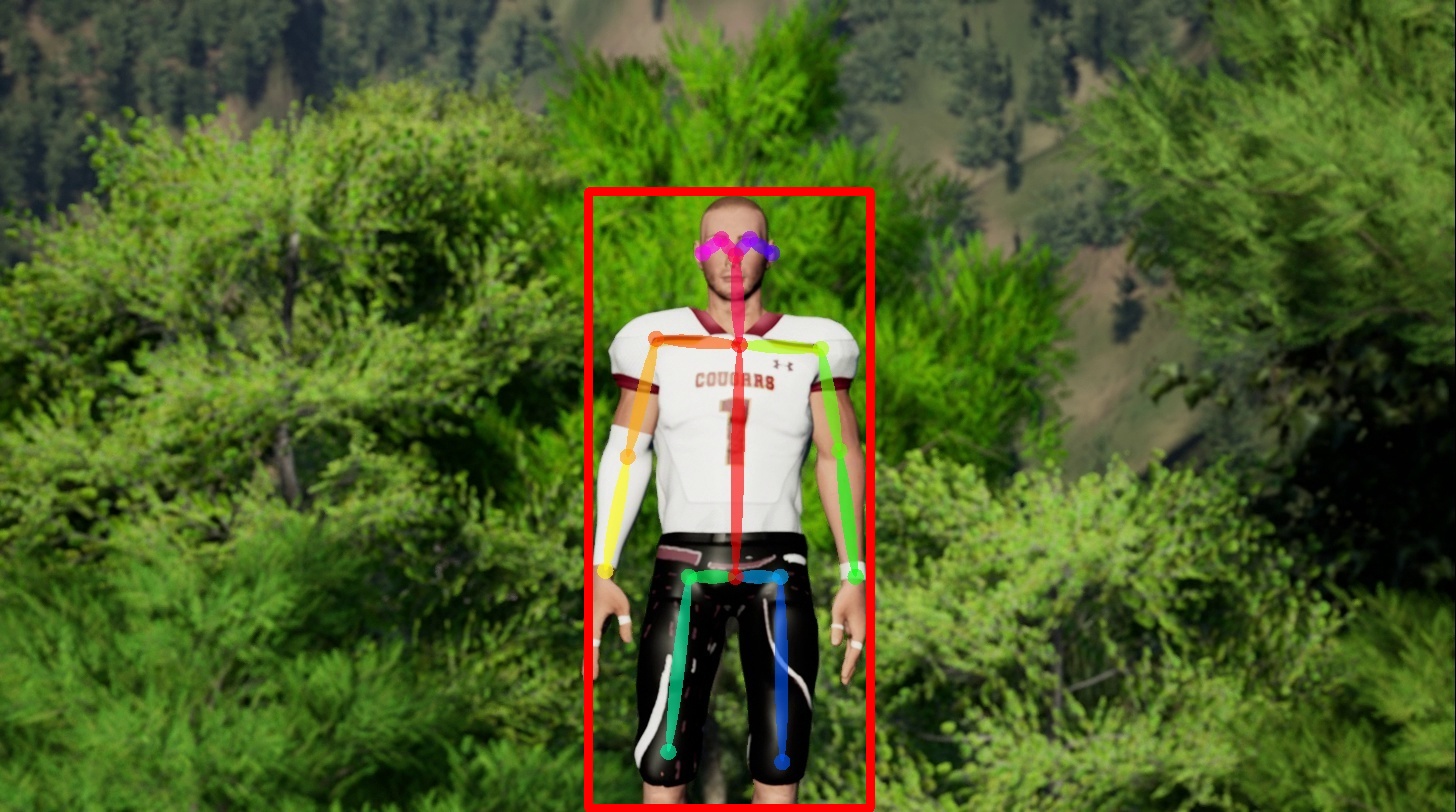}
    & \includegraphics[width=0.27\linewidth]{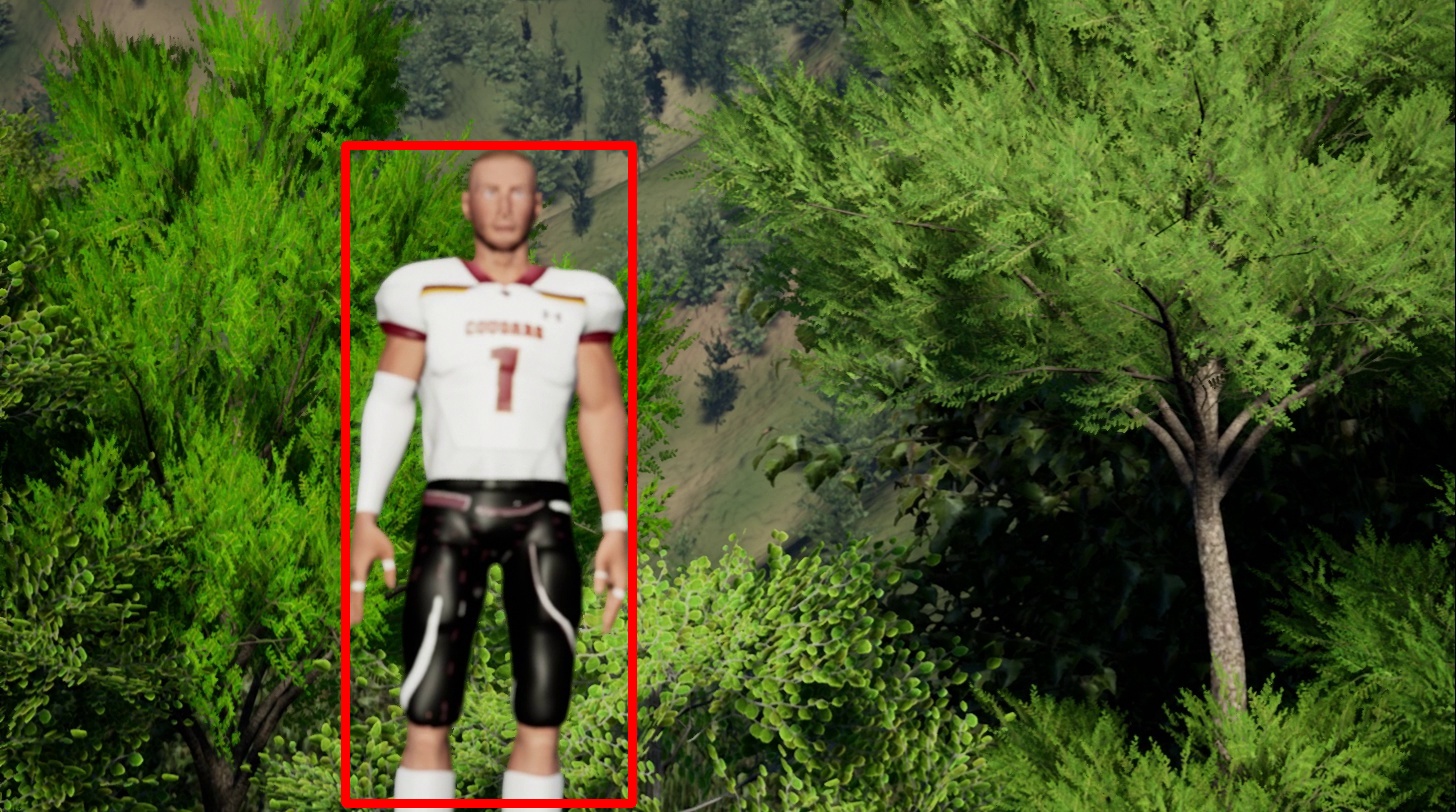}
    & \includegraphics[width=0.27\linewidth]{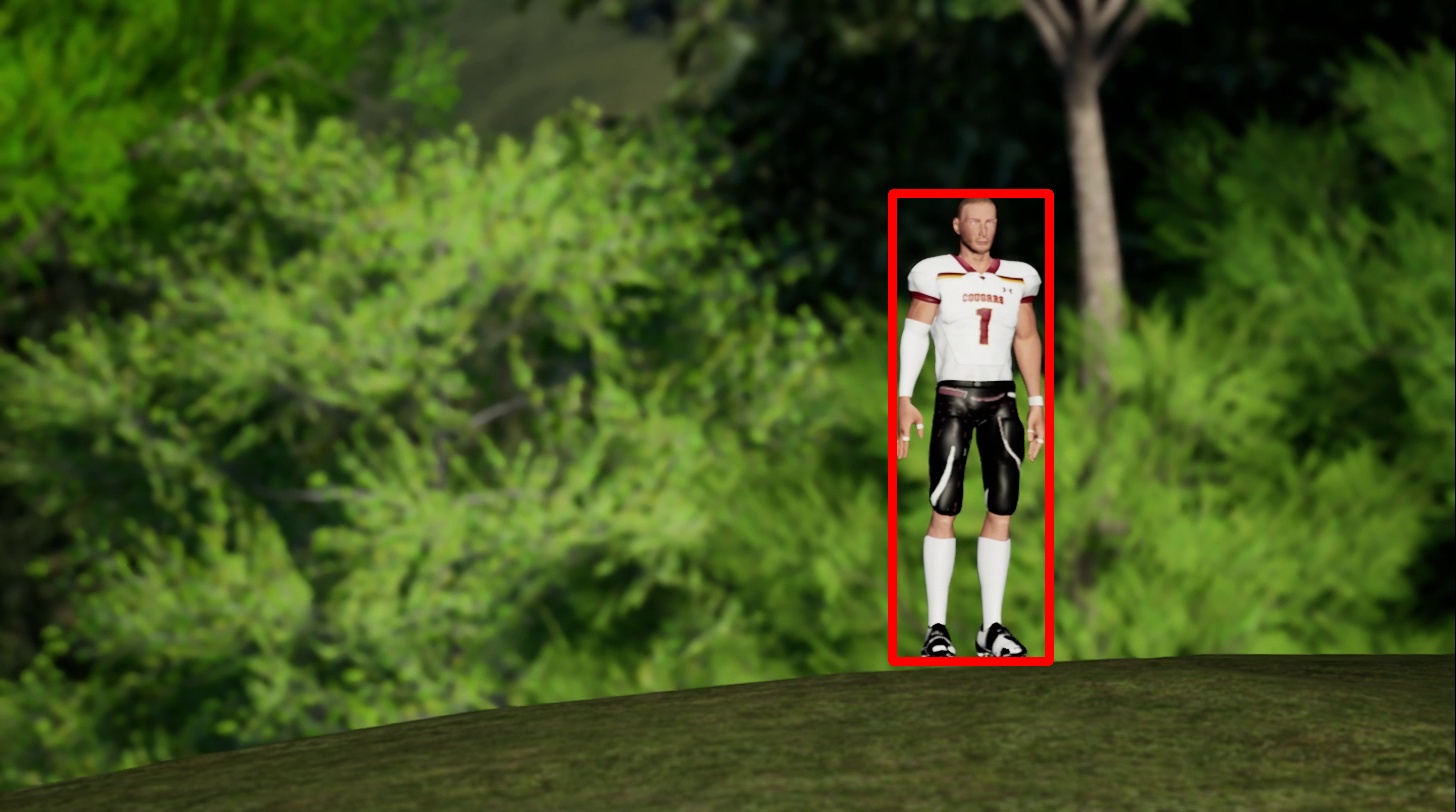}
    
\end{tabular}
\caption{\textbf {Style Transfer from S2 to T2 and T3.} 
First row: sampled frames 
from the source ordered in the time from left to right. Second and third rows: corresponding frames from the output sequences in T2 and T3.} 
\label{fig:exp_comparing_sim}
\end{figure}

This subsection shows the behavior of the whole pipeline, including CineMPC.
Each source sequence is transferred to two different scenarios: S1 to T1 and T3, and S2 to T2 and T3.
T1 and T2 are similar to S1 and S2 whereas T3 
is significantly different to both sources. 
We obtain qualitative and quantitative results by comparing the output of CineTransfer when the framework extracts the style features for both the source and output sequences. 
Figs.~\ref{fig:exp_comparing_real} and  \ref{fig:exp_comparing_sim}, and supplementary video{\textsuperscript{\footnotesize\ref{footnote_1}}} demonstrate qualitatively how CineTransfer is able to reproduce the source style. 

\subsubsection{\textbf{Transferring the focus}} 
\begin{figure}[!ht]
\centering
\begin{tabular}{cc}
    \includegraphics[width=0.43\linewidth, height=1.8cm]{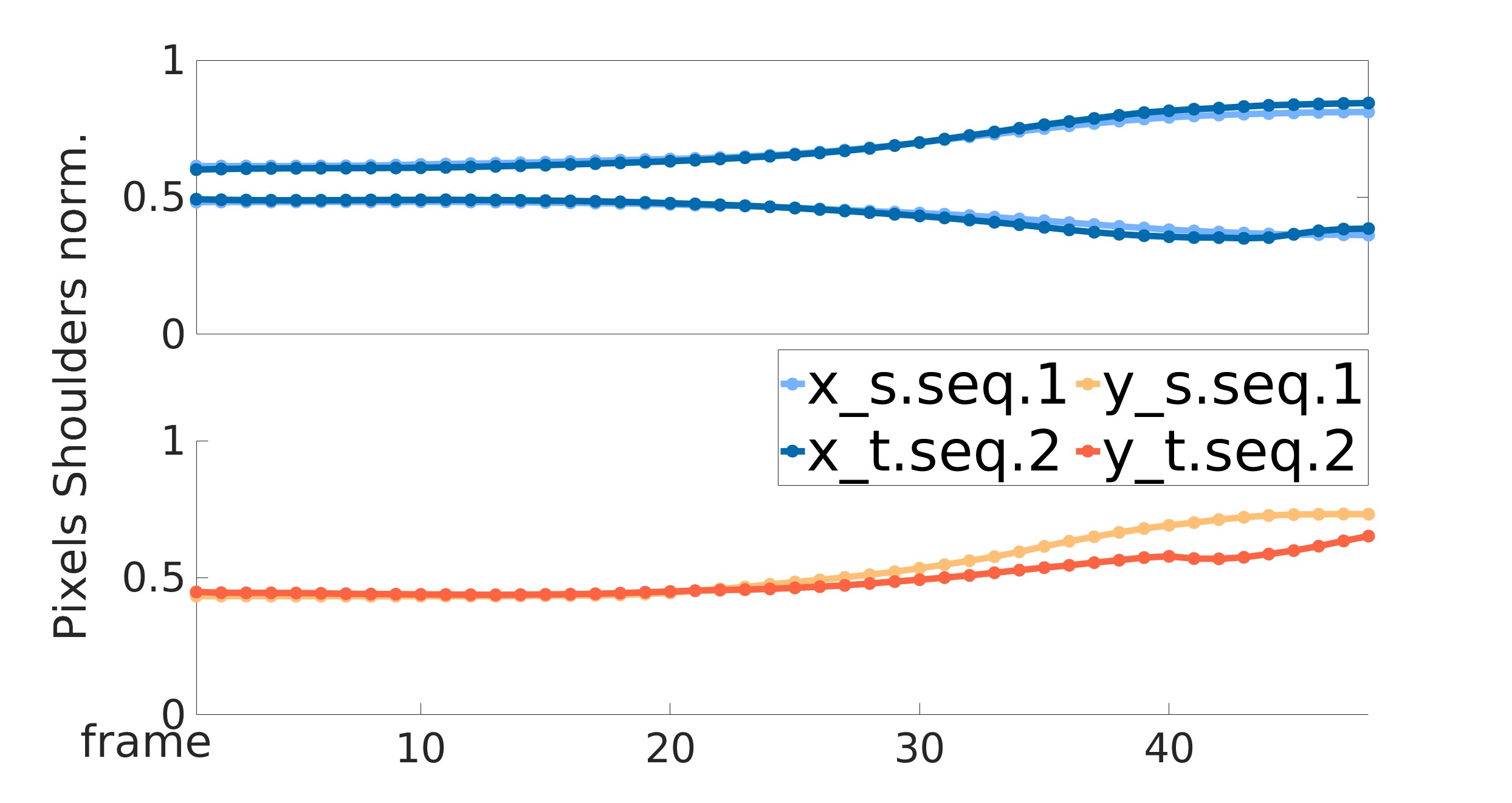} 
    &
    \includegraphics[width=0.43\linewidth, height=1.8cm]{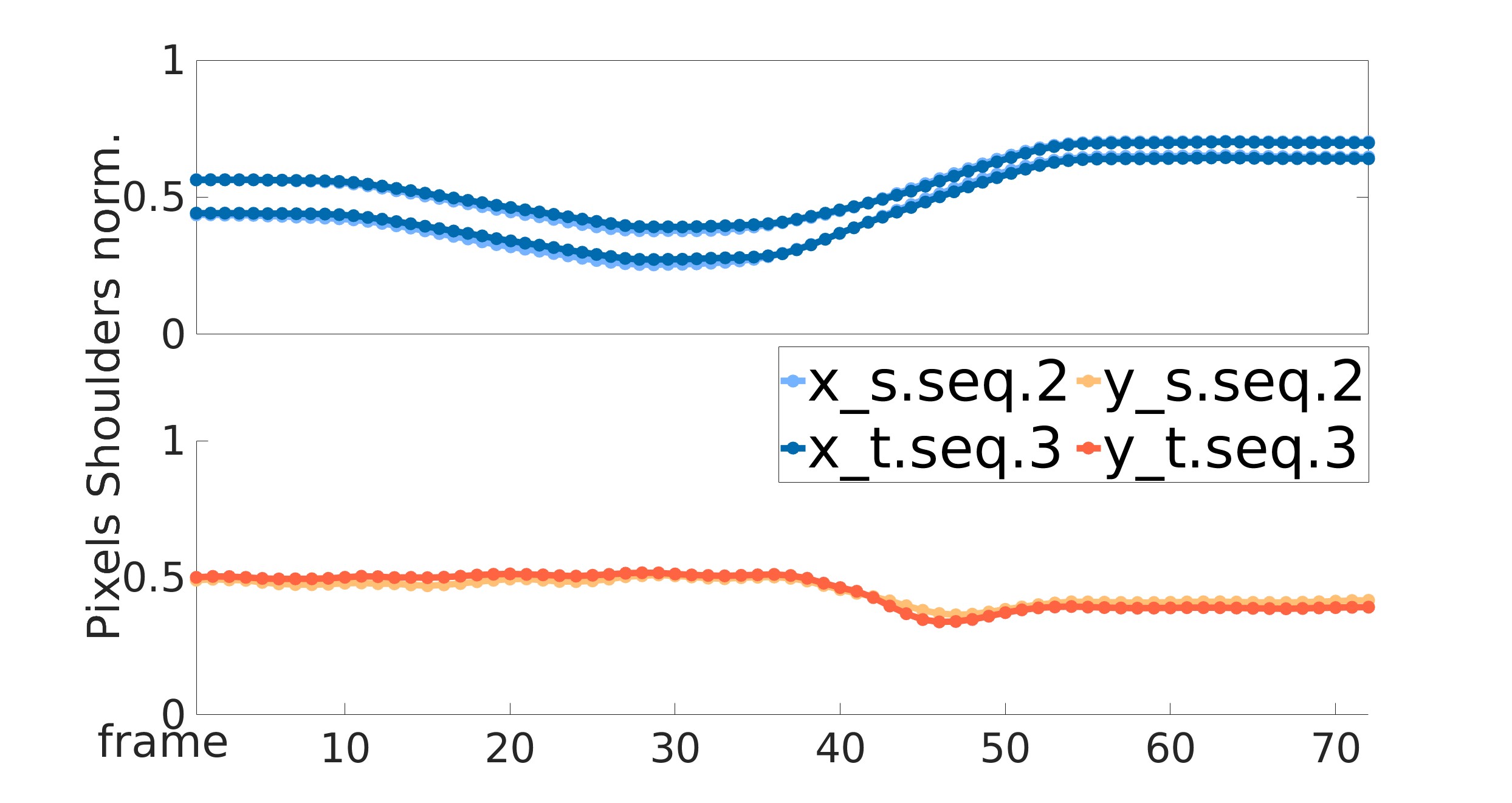} 
   \\
    \includegraphics[width=0.43\linewidth, height=1.8cm]{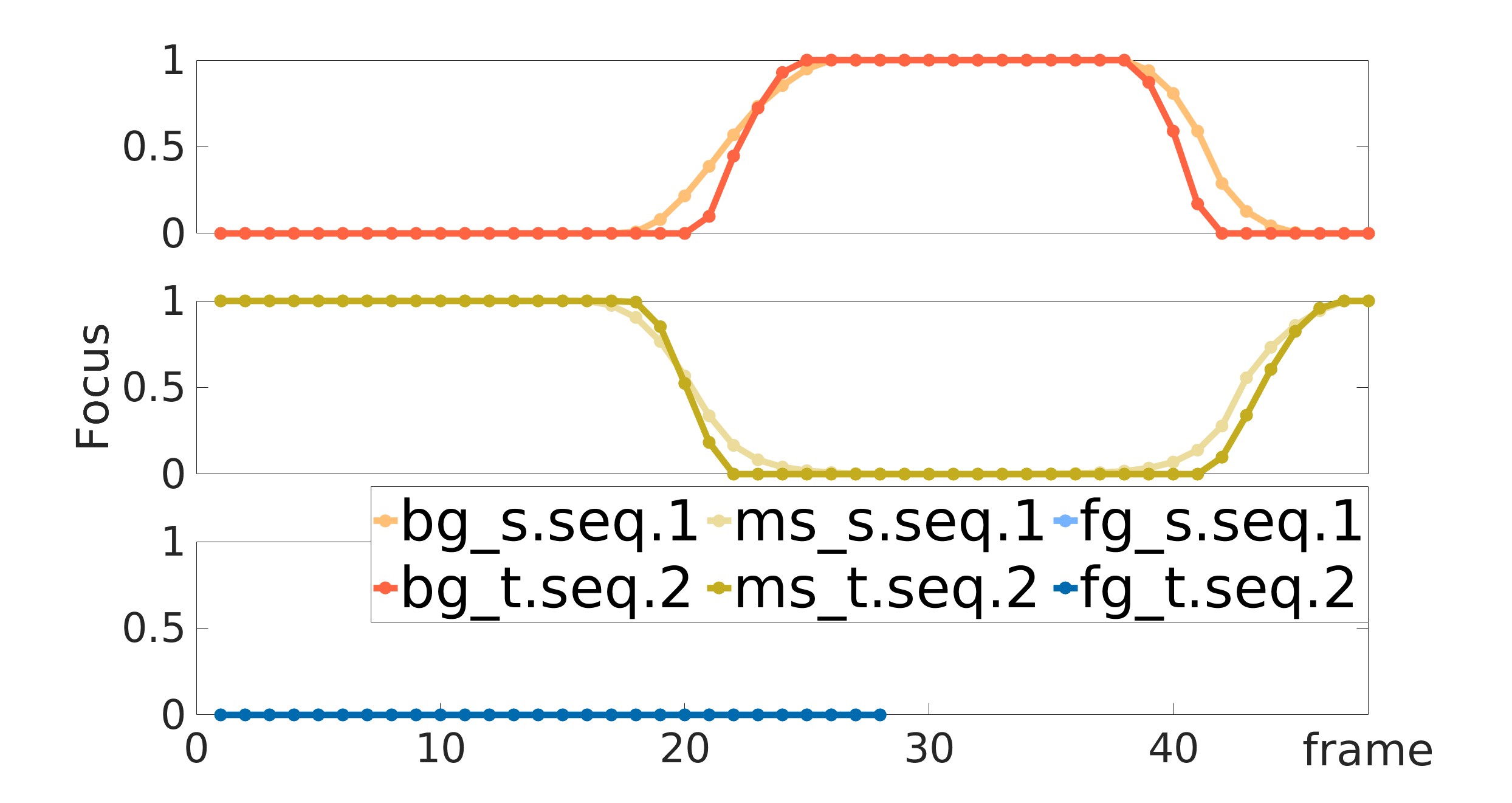} 
    & 
    \includegraphics[width=0.45\linewidth,
    height=1.8cm]{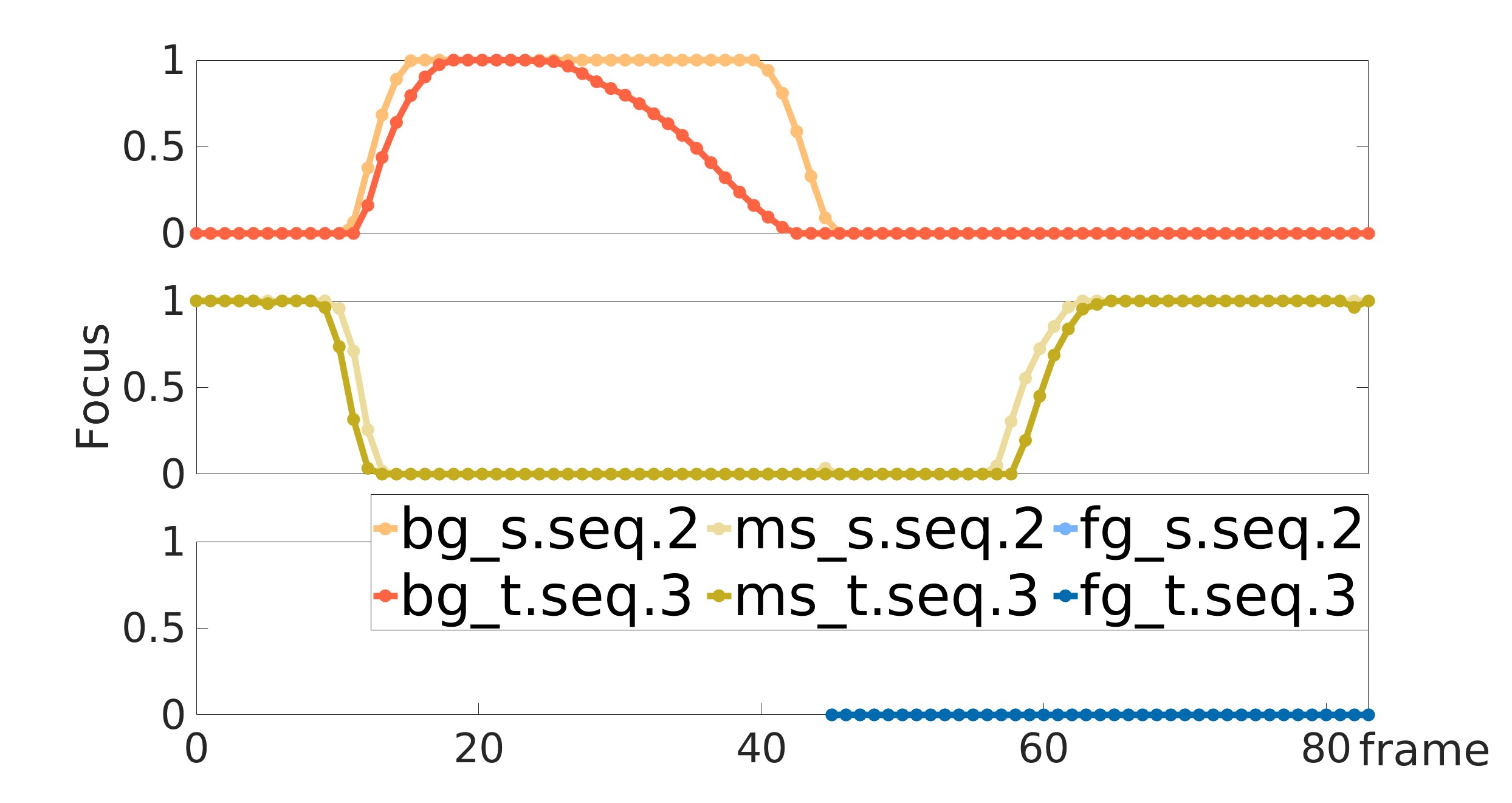}   \\
\end{tabular}
\caption{\textbf {Style Transfer. Quantitative results}. Output of the Automatic Video Style Extraction module for the source (light colors) and target sequences (dark colors). \textit{(First row)} Evolution of the image position of the joints controlled: shoulders of the main subject. Blue and orange represent horizontal and vertical components, respectively. \textit{(Second row)} Evolution of the percentage of focused pixels. Orange is the background, yellow is the subject, and blue is the foreground, only present when the subject is not shown in the bottom of the screen. '1' is focused and '0' is not focused. }
\label{fig:exp_comparing_quant}
\end{figure}

There are constant changes in the focus of the scene along the sequences. At the beginning of S1, the focus is on the main subject. 
Then, the focus goes to the background, and the subject appears out of focus. 
At the end of the sequence, the focus goes back to the subject, showing the background out of focus. 
Figure \ref{fig:exp_comparing_real} shows these frames.
We can observe similar behavior for the second source sequence, in the comparison depicted in Fig. \ref{fig:exp_comparing_sim}. 
The second row of Figure \ref{fig:exp_comparing_quant} displays the quantitative results. 

\subsubsection{\textbf{Transferring the framing of the main subject}} 
The framing of the subject changes significantly along the sequences. To transfer this feature, we control the image position of shoulders and hips. The body joints and bounding box of the subject are depicted in some frames of Figs. \ref{fig:exp_comparing_real} and  \ref{fig:exp_comparing_sim} for qualitative comparison.  
The first row of Fig. \ref{fig:exp_comparing_quant} shows the positions of subject's shoulders in source and output sequences for quantitative comparison. 
The proximity of the plots confirms a proper transfer of the framing configuration.

\section{Conclusions}
\label{sec_conclusions}
This paper presents CineTransfer, a framework that enables the extraction of the style from a source sequence and generates the commands for a mobile robot to record a new sequence reproducing the style in a new target scenario. This work builds upon CineMPC, a solution to control a camera and achieve different artistic and technical recordings. 
CineTransfer significantly increases the usability of CineMPC by enabling a much more intuitive input to the platform to define the style: a sample video. 
CineTransfer reproduces the depth of field and framing effects from the source sequence using existing extraction algorithms together with lightweight optimization without the need of extensive training datasets.
Evaluation considering both synthetic and real source sequences has demonstrated the capabilities of the framework in a photorealistic simulation environment.
Future work will explore experimentation with real mobile robots as well as the addition of other style features.

\balance
\bibliographystyle{IEEEtran}
\bibliography{references}

\end{document}